\documentclass{article}

\usepackage[preprint]{neurips_2026}

\usepackage{algorithm}
\usepackage{algpseudocode}
\usepackage[utf8]{inputenc} 
\usepackage[T1]{fontenc}    
\usepackage{hyperref}       
\usepackage{url}            
\usepackage{booktabs}       
\usepackage{amsmath}        
\usepackage{amsfonts}       
\usepackage{amssymb}        
\usepackage{nicefrac}       
\usepackage{microtype}      
\usepackage{xcolor}         
\usepackage{graphicx}
\usepackage{float}
\usepackage{caption}
\usepackage{listings}
\usepackage[capitalise,nameinlink]{cleveref}

\hypersetup{
  pdftitle={LEVI: Stronger Search Architectures Can Substitute for Larger LLMs in Evolutionary Search},
  pdfauthor={Temoor Tanveer},
}

\title{LEVI: Stronger Search Architectures Can Substitute for Larger LLMs in Evolutionary Search}

\author{%
  Temoor Tanveer \\
  Independent Researcher \\
  \texttt{ttanveer@alumni.cmu.edu} \\
}

\begin{document}

\maketitle

\begin{abstract}
LLM-guided evolutionary methods such as AlphaEvolve have proven effective in domains like math, systems research, and algorithmic discovery, but their reliance on frontier models makes each run expensive. We argue this is largely an artifact of how existing frameworks allocate search: archives that fail to preserve solution diversity force compensation through stronger mutation models; blind model use spends frontier dollars on local edits a smaller model could handle; and full-set evaluation wastes rollouts on redundant examples. We introduce LEVI, a harness-first evolutionary framework built on the bet that \emph{stronger search architectures can substitute for or even outperform larger LLMs in evolutionary search}. LEVI improves on three core components of evolutionary search: a solution database that establishes diversity from the beginning, and then maintains it throughout the run; a smarter mutation router that plays into the strengths of large and small LLMs; and a rank-preserving proxy benchmark for rollout-heavy settings. Across systems-research benchmarks LEVI attains the highest score on a budget 3.3--6.7$\times$ smaller than the published frontier-model runs of existing frameworks like ShinkaEvolve, GEPA, and AdaEvolve; on one problem, LEVI matches the existing best at a 35$\times$ lower cost. On prompt optimization, LEVI matches or exceeds GEPA at less than half of its rollout budget on four different benchmarks. LEVI is available as an open-source framework at \url{https://github.com/ttanv/levi}.
\end{abstract}

\noindent\begin{minipage}{\textwidth}
\centering
\includegraphics[width=0.85\textwidth]{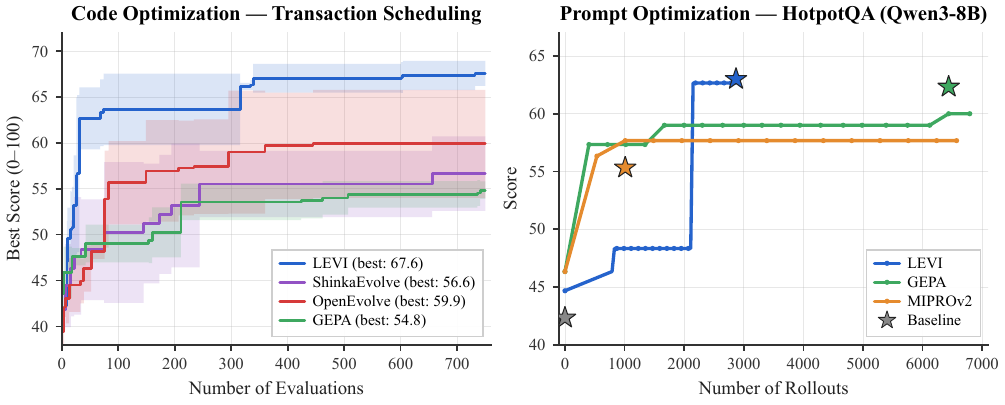}
\captionof{figure}{LEVI demonstrates that stronger search architecture can outperform larger-budget optimization. Across both code and prompt optimization, LEVI reaches higher final performance while using a fraction of the evaluation budget. Left: on transaction-scheduling code optimization, LEVI exceeds every baseline’s final score within the first $\sim$50 evaluations (15x sample efficiency). Right: on HotpotQA prompt optimization with Qwen3-8B, LEVI outperforms GEPA while using fewer than half as many rollouts, reaching its best result with $\sim$2.75K rollouts versus GEPA’s $\sim$6.87K.}
\label{fig:frontpage-code-prompt}
\end{minipage}

\section{Introduction}
\label{intro}

LLM-based evolutionary methods like AlphaEvolve \cite{novikov2025alphaevolve} have yielded promising results in different \textit{easy-to-verify, hard-to-search} domains, including math \citep{romera-paredes_mathematical_2024, georgiev2025mathematical}, code optimization \citep{suwandi2026adaptive}, heuristic design \citep{liu2024evolutionheuristicsefficientautomatic}, systems optimization \citep{cheng2025letbarbariansinai, zhao2025automatedllmspeedrunningbenchmark}, and prompt optimization \citep{guo2024connecting, gepa2025}. In this paradigm, the user defines the problem and a scoring function, while an evolutionary loop uses an LLM to mutate candidate solutions, evaluates them, and maintains a solution database of promising candidates \citep{romera-paredes_mathematical_2024,novikov2025alphaevolve}. 

Despite the demonstrated power of the paradigm, existing methods remain costly. Each result can require hundreds or thousands of calls to expensive frontier LLMs. This has been observed in recent works \citep{lange2025shinka, openevolve, cemri2026adaevolveadaptivellmdriven, liu2026evoxmetaevolutionautomateddiscovery}, where each individual run can cost up to $\$30$ and often relies on frontier LLMs such as the latest Claude Opus or Gemini Pro. These costs raise the barrier of entry for many researchers and limit the potential of the paradigm. Notably, this reliance on frontier models is not obviously inherent to LLM-guided evolution: the original FunSearch produced novel mathematical results, including the cap set improvement, without using a frontier scale LLM ~\citep{romera-paredes_mathematical_2024} . We argue much of the current cost comes from over-relying on larger models while under-investing in search architecture. We introduce LEVI, a harness-first evolutionary framework that outperforms other methods at a fraction of the cost by making the following bet: \textit{stronger search architectures can substitute for or even outperform larger LLMs in evolutionary search.} LEVI follows this by improving the search loop along three dimensions, each targeting a different source of inefficiency. 

First, on the per-evaluation axis, LEVI improves the solution database (the component responsible for maintaining diverse solutions) through diverse initialization paired with flexible, input and output-based behavioral dimensions. Rather than seeding from a single convergent-prone solution, LEVI spends a small fraction of the budget producing an explicitly diverse initial set (including weaker but structurally distinct solutions), and uses those same seeds to calibrate a CVT-MAP-Elites archive whose cells correspond to genuinely distinct algorithmic identities. The archive's robust diversity dimensions then preserve this diversity throughout the search.

Second, on the mutations per dollar axis, LEVI employs role-aware mutations: routing most ordinary mutation calls to smaller models while reserving larger models for rare paradigm shifts. The database and the router compose: strong archival diversity gives small models robustness to their own noise, while paradigm shifts from larger models inject genuinely new families into the archive.

Third, on the per-rollout axis, LEVI selects a representative proxy benchmark for settings where evaluation cost dominates; most notably rollout-heavy prompt optimization. The proxy is chosen to preserve the ranking signal of the full benchmark rather than every individual score, since evolutionary selection only depends on choosing better candidates over worse ones.

Across seven systems-research benchmarks, LEVI attains the best score on six of seven problems at 3.3–6.7× lower cost than published frontier-model runs of GEPA, OpenEvolve, ShinkaEvolve, AdaEvolve, and EvoX. Under matched-budget controlled comparison with the same model, LEVI shows up to 15x sample efficiency over other methods and continues improving where baselines stagnate. On four prompt-optimization benchmarks, LEVI exceeds GEPA on aggregate at less than half of its rollout budget. Ablations isolate the contribution of each component.


\section{Related Work}
\label{related_work}


\paragraph{LLM-guided evolutionary frameworks.}
Several open-source frameworks instantiate the FunSearch~\citep{romera-paredes_mathematical_2024}
and AlphaEvolve~\citep{novikov2025alphaevolve} paradigm with different design choices,
but still leave cost tied to frontier models, full validation, or post-hoc diversity
repair. OpenEvolve~\citep{openevolve} combines island-based evolution with
low-dimensional MAP-Elites~\citep{mouret2015illuminating} archiving; ShinkaEvolve~\citep{lange2025shinka}
adds weighted archive sampling, novelty filters, LLM novelty judges, and adaptive
model selection; AdaEvolve~\citep{cemri2026adaevolveadaptivellmdriven} adapts
exploration intensity using accumulated improvement; and EvoX~\citep{liu2026evoxmetaevolutionautomateddiscovery}
treats the search strategy itself as an evolvable object. GEPA~\citep{gepa2025}
targets prompt optimization, using natural-language reflection to mutate prompts
and maintaining diversity through per-instance Pareto fronts rather than an explicit
archive. LEVI differs by decoupling all three budget axes identified in \Cref{intro}:
a CVT-MAP-Elites archive preserves diversity by construction, role-aware routing
conditions model choice on mutation type, and proxy-benchmark selection reduces
rollout cost in validation-heavy settings.

\paragraph{Quality-diversity optimization.}
LEVI's archive design draws directly on the quality-diversity (QD) lineage. 
MAP-Elites~\citep{mouret2015illuminating} introduced the idea of partitioning 
solutions into a grid of behavioral cells, each holding the best solution to 
land there; CVT-MAP-Elites~\citep{vassiliades2017usingcentroidalvoronoitessellations} 
replaced the regular grid with a centroidal Voronoi tessellation, which scales 
gracefully as the number of behavioral dimensions grows and avoids the exponential 
cell-count explosion of the original. LEVI inherits this construction and adds 
two practitioner-facing pieces: a bootstrapped seed pass that doubles as 
descriptor calibration, and a mixed input/output descriptor family that lets 
the archive express algorithmic identity (e.g. via AST features) and behavioral 
profile (e.g. per-instance scores) simultaneously.

\section{LEVI: Cost-Efficient Harness-First Evolution}
\label{levi}

\begin{figure}[t]
  \centering
  \includegraphics[width=\textwidth]{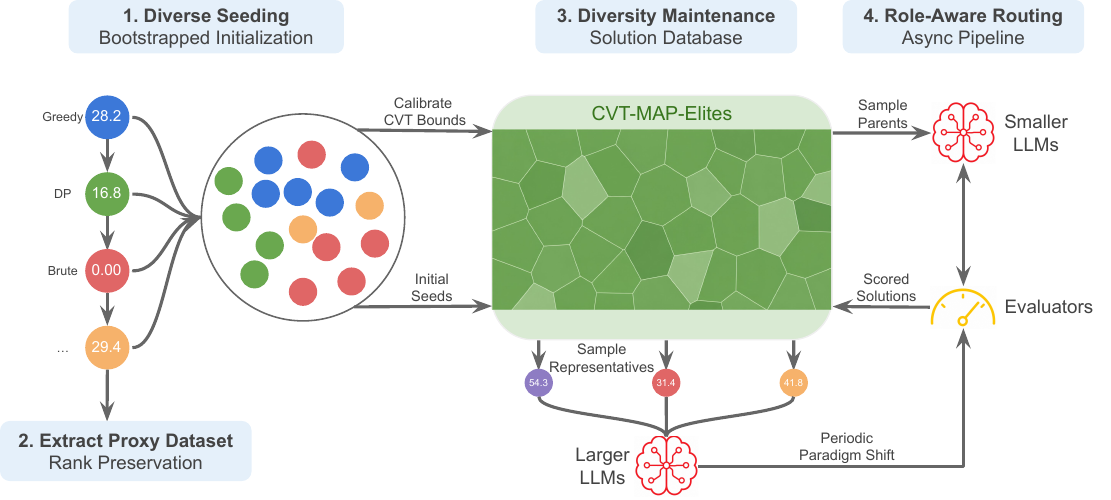}
  \caption{
LEVI uses one bootstrapped seed pass to initialize solution families, calibrate the
CVT-MAP-Elites database, and construct proxy benchmarks. During search, the archive
feeds an asynchronous mutation--evaluation loop: most mutations use a small LLM for
local refinement, while periodic paradigm-shift calls use a stronger LLM to propose
structurally new candidates.
}
  \label{fig:system_overview}
\end{figure}

We introduce LEVI, a harness-first evolutionary framework for easy-to-verify, hard-to-search
optimization problems. Given a scoring function and a fixed budget, LEVI improves the evolutionary loop along three cost axes. First, a CVT-MAP-Elites solution database converts evaluations into search progress by
preserving diversity (\Cref{seeding_archive}). Second, a role-aware LLM router reduces
mutation cost by assigning local edits to small models and reserving stronger models for rare
paradigm shifts (\Cref{llm_routing}). Third, for rollout-heavy settings, a proxy benchmark
concentrates evaluation on examples that preserve the ranking signal needed for selection
(\Cref{sec:proxy_benchmark}). \Cref{fig:system_overview} summarizes the full loop;
\Cref{alg:levi-phase1,alg:levi-phase2} specify the seeding/calibration phase and the
asynchronous evolutionary loop respectively, and the prompt templates used at each stage are
reproduced in \Cref{app:prompts}.

\textbf{Orchestration loop.}
LEVI follows an asynchronous AlphaEvolve-style loop. The solution database stores evaluated
candidates; samplers draw parents from it; the router sends mutation requests to the relevant
LLMs; and the resulting candidates are evaluated in parallel before being inserted back into the
database. By default, parents are sampled with a softmax over scores, with worker-specific
temperatures to balance exploitation and exploration across the parallel worker pool.

\textbf{Solution database.}
The solution database maintains high-scoring candidates while keeping them spread across the
problem's behavioral space. This separation is important: exploration and exploitation are handled
by the sampler, while diversity is enforced by the archive itself. When the archive collapses into a
small number of similar solutions, mutation has little structural variation to build on, and prior
systems often compensate with frontier-model calls. LEVI instead focuses on preserving diversity at insertion time.

Concretely, LEVI uses a CVT-MAP-Elites archive
\citep{vassiliades2017usingcentroidalvoronoitessellations}. Each candidate is mapped to a
behavioral descriptor: a tuple of input-side or output-side features that characterize the candidate's
algorithmic identity. The descriptor is assigned to the nearest centroid in a Voronoi tessellation, and
each cell stores only the highest-scoring candidate mapped to it. Compared with regular-grid
MAP-Elites, CVT-MAP-Elites scales to higher-dimensional descriptor spaces without exponential
cell growth, allowing LEVI to combine richer structural and behavioral signals. Descriptor values
are normalized with online z-score statistics using Welford's algorithm followed by a sigmoid
transform, so high-variance dimensions do not dominate the geometry.

\begin{figure}[t]
\noindent
\begin{minipage}[t]{0.49\textwidth}
\begin{algorithm}[H]
\footnotesize
\caption{LEVI: Phase 1 (seed \& calibrate)}
\label{alg:levi-phase1}
\begin{algorithmic}[1]
\Require
\Statex \quad scoring function $f$
\Statex \quad large / small LLMs $M_l, M_s$
\Statex \quad seed count $n_s$, variants per seed $n_v$
\Statex \quad CVT cell count $K$
\Statex \quad (opt.) discovery set $\mathcal{D}$, proxy size $K_{\mathrm{proxy}}$
\State $S \gets \emptyset$
\For{$i = 1, \dots, n_s$}
  \State $s_i \gets M_l.\textsc{DiverseSeed}(S)$ 
  \State $S \gets S \cup \{s_i\}$
\EndFor
\State $V \gets \bigcup_{s \in S} M_s.\textsc{Variants}(s, n_v)$
\State $X_0 \gets S \cup V$;\ evaluate $f$ on $X_0$
\If{$\mathcal{D} \neq \emptyset$} \Comment{proxy benchmark}
  \State $C[i,j] \gets f_j(s_i),\ s_i \in S,\ x_j \in \mathcal{D}$
  \State $\mathcal{D}_{\mathrm{proxy}} \gets \textsc{GreedyColSubset}(C, K_{\mathrm{proxy}})$
  \State restrict $f$ to $\mathcal{D}_{\mathrm{proxy}}$
\EndIf
\State $(\mu, \sigma) \gets \textsc{Welford}(\varphi(X_0))$
\State $\mathcal{C} \gets \textsc{CVT}(\hat\varphi(X_0), K)$
\State $A \gets$ empty archive over $\mathcal{C}$
\For{$x \in X_0$} \State \textsc{TryInsert}$(x)$ \EndFor
\end{algorithmic}
\end{algorithm}
\end{minipage}\hfill
\begin{minipage}[t]{0.49\textwidth}
\begin{algorithm}[H]
\footnotesize
\caption{LEVI: Phase 2 (evolutionary loop)}
\label{alg:levi-phase2}
\begin{algorithmic}[1]
\Require archive $A$ from Phase 1; budget $B$; period $\tau$
\State $t \gets |X_0|$
\While{$\mathrm{cost}(t) < B$} \Comment{$W$ workers, async}
  \If{$\textsc{Trigger}(t)$} \Comment{every $\tau$ evals}
    \State $G \gets \textsc{KMeans}(\textsc{Occupied}(A), k)$
    \State $R \gets$ top elite of each cluster in $G$
    \State $y \gets M_l.\textsc{ParadigmShift}(R)$ 
    \State eval $f(y)$;\ \textsc{TryInsert}$(y)$
  \Else
    \State parent $\sim \exp(f/T_w)$ over $\textsc{Elites}(A)$
    \State $y \gets M_s.\textsc{Mutate}(\mathrm{parent})$ 
    \State eval $f(y)$;\ \textsc{TryInsert}$(y)$
  \EndIf
  \State $t \gets t + 1$
\EndWhile
\State \Return $\arg\max_{x \in \textsc{Elites}(A)} f(x)$
\Statex
\Procedure{TryInsert}{$x$}
  \State update $(\mu, \sigma)$ on $\varphi(x)$ \Comment{Welford}
  \State $c^\star \gets \arg\min_{c \in \mathcal{C}} \|\hat\varphi(x) - c\|$
  \If{$A[c^\star] = \varnothing$ \textbf{or} $f(x) > f(A[c^\star])$}
    \State $A[c^\star] \gets x$
  \EndIf
\EndProcedure
\end{algorithmic}
\end{algorithm}
\end{minipage}
\end{figure}

\subsection{Seeding and Calibrating the Archive}
\label{seeding_archive}

A useful archive must establish diversity early and preserve it throughout 
search. Most existing frameworks begin from a single seed program 
\citep{openevolve, lange2025shinka, cemri2026adaevolveadaptivellmdriven, gepa2025}. 
This makes the early search spend budget escaping one basin of attraction, 
while later diversity is recovered through ad-hoc mechanisms such as 
frontier-model calls, islands, novelty rejection, embedding filters, or 
LLM judges.

LEVI makes a small upfront investment instead. During initialization, it asks the LLM to generate a
sequence of structurally diverse seed solutions. Each seed is conditioned on previous attempts and
explicitly instructed to differ from them; failures and error messages are included as feedback so the
next seed avoids repeated dead ends. We retain seeds even when they score poorly, because
weak-but-distinct candidates expand the observed descriptor range and provide footholds in regions
that local mutation may not reach. Seeds can also be fanned out into variants to populate additional
archive cells without repeating the full diversity-generation step.

The same seed pass calibrates the archive. For each descriptor dimension, LEVI estimates the seed
distribution, maps values through a z-score and sigmoid transform into $[0,1]$, and uses the resulting
coordinates to place candidates in the CVT tessellation. As new candidates arrive, the running
statistics are updated online, allowing the descriptor geometry to adapt as new behavioral regimes
appear.

LEVI supports two descriptor families. \emph{Input-side} descriptors are computed directly from the
solution, including code length and AST features such as cyclomatic complexity, loop count, and
operator count. These are useful when behavioral outputs are sparse or evaluated on few instances.
\emph{Output-side} descriptors are computed from execution behavior, such as runtime or
per-instance score profiles over a dataset. These expose trade-offs between examples and are
especially useful for prompt optimization. Existing frameworks commit to one family---OpenEvolve and ShinkaEvolve to input-side features \citep{openevolve, lange2025shinka}, GEPA implicitly to output-side ones via its per-instance Pareto front \citep{gepa2025}---whereas LEVI's archive accepts any mix, chosen to fit the structure of the problem.

\subsection{Role-Aware LLM Routing}
\label{llm_routing}

Mutation calls look uniform from the outside---each one takes a parent
and returns a child---but the work they do varies enormously. Most are
local refinements: tweak a constant, swap a data structure, inline a
helper. A few are structural rewrites that change the algorithm's
high-level shape. Existing frameworks often pay frontier-model prices
for both, implicitly assuming that mutation quality scales with model
scale across the board. We argue these two operations have different
requirements. Local refinement rewards speed and volume, where smaller
models are sufficient; structural rewrites reward the broader prior
coverage that comes with scale. Since the cost gap between model classes
is large, and local edits dominate the call distribution, aligning model
choice with mutation role is a direct source of dollar efficiency.

LEVI implements this split with two routes. The \emph{refinement route}
sends the bulk of mutation calls---roughly $90\%$ in our experiments---to
a small model such as Qwen3-30B-A3B \citep{qwen3technicalreport}. Parents are drawn from the archive
using the standard softmax-over-scores sampler, and the model is asked
for a single targeted edit. Smaller models can collapse onto common
patterns, but the CVT-MAP-Elites archive absorbs this failure mode:
near-duplicate outputs map to occupied cells and are discarded. Collapse
pressure is therefore converted into harmless redundancy rather than
archive drift.

The \emph{paradigm-shift route} is where the larger model earns its
cost. Instead of asking it for anothers edit on a single strong parent,
LEVI samples high-scoring representatives from distinct, well-separated
archive cells (enabled by the CVT-MAP-Elites's naturally separable regions) and passes them together as context. The large model is
then asked to propose a candidate that does not belong to any of those
families. In this route, the model operates on the structure of the
population rather than on one parent: its role is to extend the frontier
of explored solution families, not merely improve an existing one. We
trigger such calls at fixed intervals and on stagnation.

The two routes are complementary. Refinement deepens the cells the
archive already occupies; paradigm shifts expand the set of cells worth
occupying. Without the small-model route, the loop spends frontier-model
budget on edits that cheap models can handle. Without the large-model
route, the search risks polishing only the families discovered during
initialization. The optimal split is problem-dependent: some tasks reward
many local refinements, while others benefit from periodic structural
jumps. The archive ties both routes together, giving the small model
robustness to its own redundancy and giving the large model a structural
view of the population that a single-parent prompt cannot.

\subsection{Proxy Benchmark Selection}
\label{sec:proxy_benchmark}

The third allocation axis appears when evaluating a candidate solution is itself expensive. 
In code-optimization tasks, the dominant cost is often the mutation call or the program evaluation. 
In prompt optimization, however, each candidate prompt may require many rollout evaluations across a validation set, and prior work spends a substantial fraction of its budget on this repeated validation step~\citep{gepa2025}. 
LEVI therefore constructs a smaller \emph{proxy benchmark} $D_{\mathrm{proxy}} \subset D$ that preserves the selection signal of the full validation set while reducing the number of rollouts required during evolution.

The proxy benchmark is built during the same initialization phase used to seed and calibrate the solution database. 
Let $p_1,\ldots,p_m$ denote the diverse seed candidates produced during initialization, and let $D=\{x_1,\ldots,x_n\}$ denote the full discovery set. 
We evaluate each seed candidate on each example in $D$ to obtain a calibration matrix
\[
    M \in \mathbb{R}^{m \times n}, \qquad M_{ij} = \mathrm{score}(p_i, x_j).
\]
This matrix estimates how candidate quality varies across examples. 
The goal is to select a subset $S \subseteq \{1,\ldots,n\}$ of size $K_{\mathrm{proxy}}$ such that the average score over $S$ induces nearly the same ranking over candidates as the average score over the full set $D$.

LEVI chooses $S$ with greedy forward selection. 
Starting from $S=\emptyset$, each step adds the example $j \notin S$ that maximizes a marginal score:
\[
\mathrm{score}(j \mid S)
=
\lambda_r R(S \cup \{j\})
+
\lambda_s A(S \cup \{j\})
-
\lambda_c C(j,S).
\]

The objective combines three terms. \emph{Rank faithfulness} $R(S)$ rewards subsets whose induced candidate ranking agrees with the full-set ranking, directly matching what evolutionary selection consumes; we measure it as the fraction of candidate pairs $(p_a, p_b)$ whose ordering under proxy scoring agrees with their ordering under full scoring, with two-sided ties counted as agreement, one-sided ties receiving partial credit, and reversals counted as disagreement. \emph{Separation} $A(S)$ rewards examples that discriminate between candidates, measured as the average normalized column standard deviation of $M$ across calibration candidates; examples on which all candidates behave similarly carry little selection signal even when they are representative. \emph{Redundancy} $C(j, S)$ penalizes adding example $j$ when its score column is highly correlated (mean absolute Pearson) with those already in $S$, since correlated examples tend to rank candidates the same way and add little new information. We use fixed weights $\lambda_r = \lambda_s = 0.5$ and $\lambda_c = 0.15$ across all benchmarks, treating rank preservation and separation as the two primary objectives and redundancy as a soft regularizer.

This procedure is general: it can be applied whenever an initial calibration set of candidates can be evaluated on a larger pool of examples.
In this paper we use it primarily for prompt optimization, where rollout cost dominates and preserving the ranking over candidate prompts is sufficient to guide evolution.
We empirically validate this design choice in \Cref{app:proxy_ablation}, where CSS+mean dominates two natural subset-selection baselines ($k$-medoids and random-subset + ridge regression) at every iso-cost contour on a $24$-prompt $\times$ $150$-problem score matrix.

\section{Evaluation}
\label{sec:eval}

For code optimization, we evaluate on systems-research benchmarks from prior work,
spanning combinatorial scheduling, databases, LLM serving, and distributed systems
\citep{cheng2025letbarbariansinai}. For prompt optimization, we rely on already established problems used by existing works \citep{gepa2025}. We provide more details pertaining to the experiments at \Cref{app:hyperparams}, \ref{app:defaults}, and \ref{app:ablation_conditions}.

\begin{table}[!ht]
\caption{Systems-research comparison against published frontier-budget results. Baseline entries are the best reported runs from prior frameworks using GPT-5 or Gemini 3.0 Pro at \$15--30 per problem; LEVI uses smaller models for most mutations and Gemini Flash 3.0 only for paradigm shifts. LEVI attains the best value on most problems at a fraction of the cost. On the LLM-SQL problem, LEVI outperforms the leading framework after using only \$0.5, lowering the cost by over 35x.}     
\centering                             
\small                         
\setlength{\tabcolsep}{4pt}
\renewcommand{\arraystretch}{1.15}        

\setlength{\tabcolsep}{0pt}
\begin{tabular*}{\linewidth}{@{\extracolsep{\fill}}lccccccc@{}}                              
\hline                           
\textbf{Framework} & \textbf{Cloudcast ($\downarrow$)} & \textbf{EPLB ($\uparrow$)} & \textbf{LLM-SQL ($\uparrow$)} & \textbf{Prism ($\uparrow$)} &
\textbf{Txn Sched ($\uparrow$)} & \textbf{Spot-M ($\uparrow$)} & \textbf{Spot-S ($\uparrow$)} \\                  
\hline                                                    
GEPA          & 645.72 & 0.1445 & 0.7134 & 26.23 & 3984.1  & 62.2 & 51.4 \\                         
OpenEvolve    & 707.82 & 0.1272 & 0.7258 & 26.24 & 4273.5  & 66.7 & 42.5 \\                           
ShinkaEvolve  & 812.74 & 0.1272 & 0.7212 & 26.26 & 4329.0  & 63.6 & 45.6 \\                           
AdaEvolve     & 637.10 & 0.1453 & 0.7750 & \textbf{26.37} & 4348.0  & ---  & ---  \\                           
EvoX          & 623.69 & 0.1453 & 0.7300 & 26.26 & 4347.83 & ---  & ---  \\                           
\textbf{LEVI} & \textbf{578.10} & \textbf{0.1523} & \textbf{0.7985} & 26.26 & \textbf{4464.29} & \textbf{72.4} & \textbf{51.7} \\                     
\hline                           
\multicolumn{8}{@{}l}{\textit{Total optimization budget (\$)}} \\  
\textit{Baseline} & \$15   & \$15   & \$20   & \$15   & \$20    & \$25  & \$30  \\                          
\textit{LEVI}     & \$4.50 & \$4.50 & \$4.50 & \$4.50 & \$12.50 & \$4.50 & \$4.50 \\                                                  
\hline                          
\end{tabular*}
\label{tab:adrs}                

\end{table}

\subsection{AI-Driven Systems Research}
We evaluate on seven tasks from the ADRS problem suite~\citep{cheng2025letbarbariansinai,mang2026frontiercs}, spanning networking, LLM serving, databases, and distributed systems. For baselines, we use the strongest published ADRS results from different frameworks (GEPA, OpenEvolve, ShinkaEvolve, AdaEvolve, EvoX) coupled with expensive budgets (up to $\mathdollar30$) and access to SOTA models like GPT-5 and Gemini 3.0 Pro \citep{liu2026evoxmetaevolutionautomateddiscovery,cemri2026adaevolveadaptivellmdriven}. On the other hand, LEVI uses $<5\mathdollar$ on most problems, while using a Qwen 30B and a MiMo-V2-Flash (309B total, 15B active parameters) \citep{coreteam2026mimov2flashtechnicalreport} for more than $90\%$ of mutations. The rest are routed to Gemini 3.0 Flash for paradigm shifts.

\paragraph{Result 1: A stronger search architecture can outperform other frameworks despite using weaker models and a smaller budget:} LEVI improves on the best published result on six of the seven native metrics: lower Cloudcast cost (578.10 vs.\ 623.69), higher EPLB quality (0.1523 vs.\ 0.1453), higher LLM-SQL quality (0.7985 vs.\ 0.7750), higher Transaction Scheduling score (4464.29 vs.\ 4348.0), and stronger Spot-M / Spot-S scores (72.4 / 51.7 vs.\ 66.7 / 51.4). Prism is the exception: all frontier-budget methods cluster within 0.14 points. On six problems LEVI spends \$4.50, giving 3.3--6.7$\times$ lower cost than the corresponding frontier-model baselines; on Transaction Scheduling, the extended \$12.50 run remains below the \$20 frontier-model budget while outperforming the existing best by a large margin.

\begin{figure}[t]
    \centering

    \includegraphics[width=0.9\textwidth]{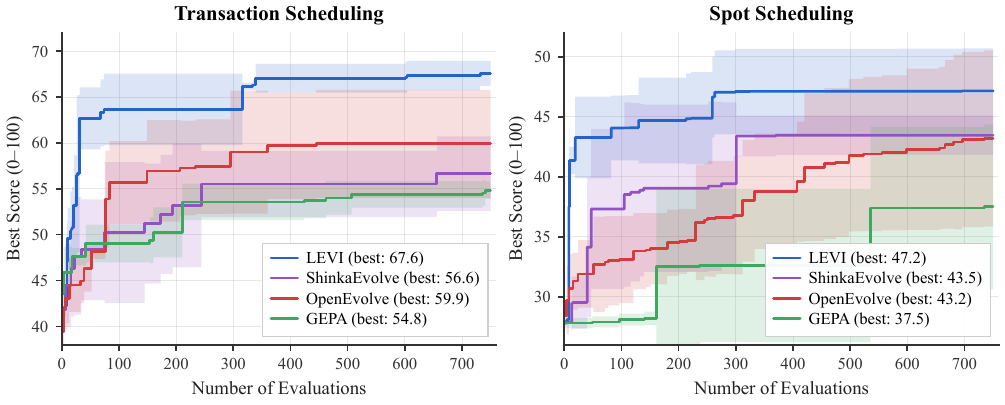}
    \caption{Controlled architecture comparison on Transaction Scheduling (left) and Spot Scheduling (right).  All frameworks use the same model (Qwen3-30B-A3B), the same evaluation budget (750 successful evaluations), and three random seeds (mean $\pm$ std shown).  LEVI reaches high performance substantially earlier on both problems and achieves the highest final score, despite identical model access.}
    \label{fig:controlled-comparison}
\end{figure}

\paragraph{Result 2: Under identical settings, seeding and calibrating the archive is highly evaluation-efficient.}
To isolate the effect of search architecture, we run LEVI, OpenEvolve, ShinkaEvolve, and GEPA under identical conditions: Qwen3-30B-A3B~\citep{qwen3technicalreport}, 750 evaluations, and three random seeds per framework. We use two complementary problems: \emph{Transaction Scheduling}, a single-output NP-hard ordering task, and \emph{Spot Scheduling}, which is scored across 1{,}080 simulations and therefore gives Pareto-style methods a richer signal.

Figure~\ref{fig:controlled-comparison} shows that LEVI reaches high scores much earlier under the same model and evaluation budget. On Spot Scheduling, LEVI reaches its near-peak score by $\sim$50 evaluations, while OpenEvolve requires over 600 evaluations to approach the same level, a roughly 12$\times$ sample-efficiency gap. On Transaction Scheduling, LEVI reaches 62 within the first 100 evaluations---a level no baseline reaches during the run---and finishes at 67.6, compared to 59.9 for OpenEvolve and 54.4 for GEPA. The baselines plateau earlier, while LEVI continues improving through the middle of the run, consistent with the calibrated archive sustaining exploration beyond the first discovered family. 

\begin{figure}[t]
  \centering
  \includegraphics[width=0.9\textwidth]{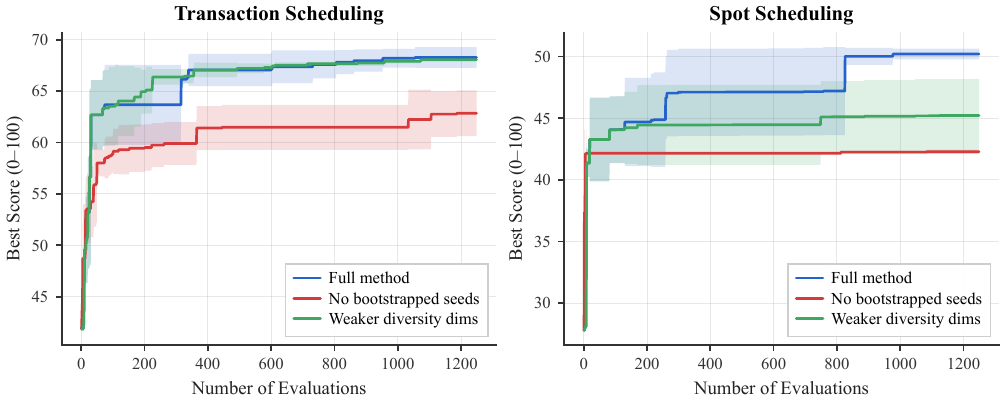}
  \caption{Ablation on the solution database, studying the impact of bootstrapped initialization and our diversity dimensions. Across the
transaction scheduling and the spot scheduling problems, bootstrapped initialization coupled with strong diversity dimensions remains the strongest
option. The former allows rapid initial improvements while the latter ensures the archive does not converge.}
  \label{fig:diversity-ablation}
\end{figure}

\paragraph{Result 3: Bootstrapped initialization consistently improves performance, while richer diversity dimensions help most when multiple algorithmic families matter.} To isolate the two design choices behind our solution database---bootstrapped seed-based initialization and the rich AST-derived diversity dimensions (\Cref{seeding_archive})---we ablate each on Transaction Scheduling and Spot Scheduling using the same Qwen3-30B-A3B model with three random seeds per variant. Detailed results in Figure~\ref{fig:diversity-ablation}. Removing bootstrapped seeds by far has the largest impact: on Transaction Scheduling the final score falls from $68.3 \pm 1.0$ to $62.8 \pm 2.2$, and on Spot Scheduling the variant stagnates near the un-optimized baseline ($42.3 \pm 0.2$ vs.\ $50.2 \pm 0.4$), essentially flat from evaluation 100 onward. The diversity-dimensions ablation is problem-dependent---weaker dimensions match the full set on Transaction Scheduling ($68.0$ vs.\ $68.3$) but lose $5$ points on Spot Scheduling ($45.2$ vs.\ $50.2$), where a broader range of viable algorithmic families need to be maintained and explored. Archive-family analysis in \Cref{app:archive-composition} confirms this pattern: stronger dimensions maintain multiple viable families on Spot Scheduling, while Transaction Scheduling remains dominated by one conflict-aware-greedy family.

\begin{figure}[t]
  \centering
  \includegraphics[width=\textwidth]{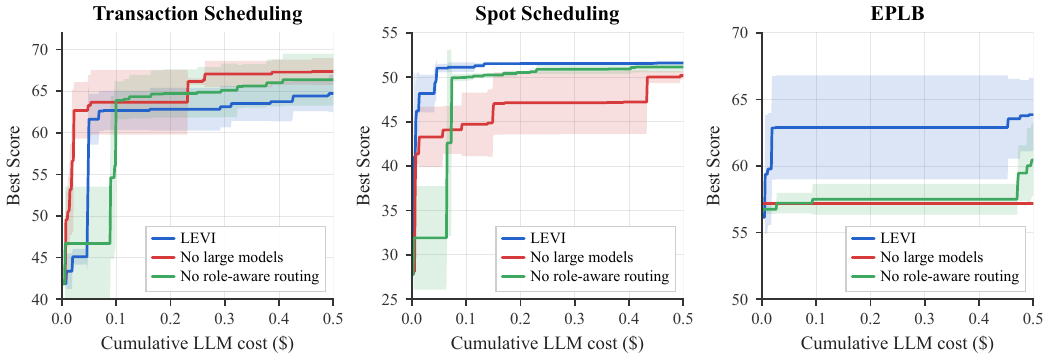}
  \caption{Ablation on the model allocation, studying the impact of using larger models and role-aware routing. For transaction scheduling, no large models yields best performance due to its local improvements heavy regime. For cloud scheduling and load-balancing, full LEVI shows clear improvements by outperforming others at 1/5th of the budget. }
  \label{fig:model_alloc-comparison}
\end{figure}

\paragraph{Result 4: Under fixed dollar budgets, model allocation matters, but larger models help only when structural jumps are needed.}
We ablate LEVI's paradigm-shift model and role-aware routing on Transaction Scheduling, Spot Scheduling, and EPLB, plotting best score against cumulative LLM cost up to \$0.50 with three random seeds per variant. LEVI routes ${\sim}90\%$ of mutations to Qwen3-30B-A3B and ${\sim}10\%$ to Gemini Flash 3 for paradigm shifts; \emph{no role-aware routing} uses the same large and small model mix but has no role distinction; and \emph{no large models} removes Gemini entirely. The results reveal three regimes. Transaction Scheduling is dominated by local refinement within one algorithmic family: all variants converge to similar scores by \$0.50, and \emph{no large models} slightly beats LEVI ($67.4$ vs.\ $64.7$). This is consistent with the harness-first thesis: when the search does not need to escape its current basin, a strong archive plus a cheap model is sufficient, and paradigm-shift calls become overhead. Spot Scheduling is intermediate: all variants reach similar final scores (${\sim}50$--$52$), but LEVI gets there earliest. EPLB is where paradigm shifts decisively matter: \emph{no large models} stagnates at $57.2 \pm 0.0$ with structurally identical elites, while LEVI reaches $63.9 \pm 2.7$ by discovering families Qwen-only never reaches. LEVI also matches the \emph{no large models} \$0.50 endpoint by roughly \$0.10 on EPLB, a $5\times$ matched-score cost advantage (Figure~\ref{fig:model_alloc-comparison}).

\subsection{Prompt Optimization}
We evaluate on four standard prompt-optimization benchmarks from the GEPA evaluation suite~\citep{gepa2025}---HotpotQA (\cite{yang2018hotpotqa}), IFBench (\cite{pyatkin2025generalizing}), Hover (\cite{jiang2020hover}), and PUPA (\cite{siyan2025papillon})---with Qwen3-8B as the task model for both candidate generation and validation rollouts, and use the train/validation/test splits provided by that suite. All methods optimize the same DSPy program (shared module structure, signatures, and evaluation harness) \cite{khattab2023dspy}, so only the prompts produced by each optimizer differ. We measure cost in total rollouts, since each candidate evaluation requires a full pass over the validation set and this dominates the per-iteration budget.

\paragraph{Result 5: Proxy benchmarks give enough signals for prompt optimization to succeed} Table~\ref{tab:levi_vs_gepa} reports the results. LEVI attains the highest aggregate score (62.02, a $+13.17$ gain over the un-optimized baseline) while spending \textit{less than half} of GEPA's rollout budget on aggregate ($2{,}191$ vs.\ $4{,}985$ rollouts averaged across the four tasks). The per-task picture is consistent: LEVI matches GEPA on HotpotQA (63.00 vs.\ 62.33) and substantially improves on IFBench (46.33 vs.\ 38.61, $+7.72$), while remaining competitive on Hover (49.00 vs.\ 52.33) and PUPA (89.73 vs.\ 91.85). Per-task rollout ratios fall in the $40$--$53\%$ range, so the efficiency advantage holds uniformly rather than coming from any single benchmark. The efficiency gain comes primarily from proxy-benchmark selection (\Cref{sec:proxy_benchmark}), which compresses each iteration to a small ranking-faithful subset while reserving full-set evaluation for late-stage selection. The same archive and routing machinery transfer from code optimization unchanged.

\begin{table}[!htbp]
\caption{Benchmark results for different optimizers evaluated on a Qwen 3 8B. LEVI performs better or competitively while using roughly half of the rollouts. On the aggregate it is the best performing optimizer while using the least rollouts.}
\label{tab:levi_vs_gepa}
\centering                          
\begin{tabular}{lcccccc}                          
\toprule                          
\textbf{Qwen3 8B} & \textbf{HotpotQA} & \textbf{IFBench} & \textbf{Hover} & \textbf{PUPA} & \textbf{Aggregate} & \textbf{Improvement} \\              
\midrule                          
Baseline & 42.33 & 36.90 & 35.33 & 80.82 & 48.85 & --- \\ 
MIPROv2  & 55.33 & 36.22 & 47.33 & 81.55 & 55.11 & +6.26 \\                          
GEPA     & 62.33 & 38.61 & \textbf{52.33} & \textbf{91.85} & 61.28 & +12.44 \\                        
LEVI     & \textbf{63.00} & \textbf{46.33} & 49.00 & 89.73 & \textbf{62.02} & \textbf{+13.17} \\                          
\midrule                          
\multicolumn{7}{l}{\textit{Total optimization budget (\# rollouts)}} \\                           
GEPA     & 6871 & 3593 & 7051 & 2426 & 4985 & --- \\                          
LEVI     & \textbf{2750} & \textbf{1870} & \textbf{2870} & \textbf{1275} & \textbf{2191} & --- \\
\bottomrule                          
\end{tabular}                                             
\end{table}

\section{Conclusion}
\label{sec:conclusion}

LEVI rests on a single bet: the practical cost of LLM-guided evolutionary search can be lowered far more by investing in the harness than by reaching for a larger mutation model. We instantiate this through three coupled components---a CVT-MAP-Elites solution database with bootstrapped, behaviorally diverse initialization; a role-aware router that sends ${\sim}90\%$ of mutations to a small open-weight model and reserves a stronger model for periodic paradigm shifts; and a rank-faithful proxy benchmark for rollout-heavy settings---each addressing one of three cost axes. Across systems-research and prompt-optimization benchmarks, LEVI improves or matches stronger-model baselines at substantially lower dollar or rollout budget, and the ablations show that the three components contribute in different regimes: seeding and descriptors preserve useful diversity, role-aware routing trades off local refinement against structural jumps, and proxy benchmarks preserve the ranking signal needed for selection.

\paragraph{Limitations and Future Work.}
LEVI reduces dollar cost by shifting much of the work from frontier-model calls to cheaper mutations and stronger search architecture, but this can require more evaluations. This trade-off is favorable when evaluations are cheap, as in most of our systems benchmarks, but may be less attractive when each evaluation takes hours, such as model training. We also do not fully study wall-clock efficiency as a separate axis, although LEVI's asynchronous design suggests room for further gains. Finally, LEVI still reserves a small fraction of calls for frontier models; an important next step is to study regimes that rely entirely on cheap open-weight models.


\begin{ack}
We gratefully acknowledge Google's TPU Research Cloud (TRC) program, whose TPU compute supported serving Qwen3-30B-A3B locally during much of the development and experimentation that informed LEVI's design.
\end{ack}

\bibliographystyle{plainnat}
\bibliography{references}


\clearpage
\appendix
\crefalias{section}{appendix}

This appendix supports reproducibility for the experiments reported in the main paper. \Cref{app:hyperparams} lists per-benchmark settings for the seven ADRS problems and the four prompt-optimization tasks. \Cref{app:defaults} lists LEVI's default configuration. \Cref{app:ablation_conditions} defines the three variants used in each ablation. \Cref{app:prompts} reproduces the prompt templates LEVI sends at each stage of the search loop. \Cref{app:pe_details} describes the punctuated-equilibrium trigger and execution flow. \Cref{app:eplb-sibp} reproduces in full the discovered EPLB elite referenced in \Cref{sec:eval}. \Cref{app:proxy_ablation} validates the proxy-benchmark selection objective against $k$-medoids and random-subset + ridge baselines.

\section{Per-benchmark experimental settings}
\label{app:hyperparams}

\Cref{tab:per-benchmark} lists the LEVI configuration used for each ADRS systems-research benchmark. All problems use the same paradigm-shift model (Gemini Flash 3) and a CVT-MAP-Elites archive with 50 centroids; what varies is the mutation-model mix, the dollar budget, and the behavioral descriptors fed into the archive. Mutation models are sampled across four softmax temperatures $\{0.3, 0.7, 1.0, 1.2\}$ as described in \Cref{app:defaults}. Score-key descriptors (when present) are per-instance score columns appended to the AST descriptor; this turns the archive into an output-side as well as input-side map.

\begin{table}[!ht]
\centering
\small
\setlength{\tabcolsep}{5pt}
\renewcommand{\arraystretch}{1.18}
\caption{ADRS benchmark semantics and native metrics. Cloudcast, Prism, and EPLB are placement/routing problems; LLM-SQL and Transaction Scheduling are reordering problems; Spot Single and Spot Multi are online scheduling policies evaluated in trace-driven simulators.}
\label{tab:adrs-benchmark-semantics}
\begin{tabular}{p{0.16\linewidth}p{0.52\linewidth}p{0.23\linewidth}}
\toprule
\textbf{Benchmark} & \textbf{Task} & \textbf{Native metric} \\
\midrule
Cloudcast ($\downarrow$) &
Multi-cloud data transfer optimization. The evolved policy builds a multicast overlay with intermediate waypoint regions, reducing egress cost relative to direct replication from a source to all destinations. &
Total cloud egress cost; lower is better. \\
EPLB ($\uparrow$) &
Expert Parallelism Load Balancer for MoE inference. The evolved policy maps expert replicas to GPUs to avoid hot devices while keeping the rebalancing computation efficient. &
Combined load-balance and runtime quality relative to the EPLB target; higher is better. \\
LLM-SQL ($\uparrow$) &
Row/column reordering for LLM inference over tabular context. The evolved heuristic orders table content to increase KV-cache prefix reuse while keeping the reordering step fast. &
Cache-hit quality with runtime penalty; higher is better. \\
Prism ($\uparrow$) &
Model placement for multi-LLM serving on a shared GPU pool under bursty heterogeneous demand. The evolved policy improves placement by reducing load hot spots, typically through local swap/move refinements over greedy placement. &
Placement quality derived from maximum KV pressure ratio; higher is better. \\
Txn Sched ($\uparrow$) &
Offline transaction scheduling. Given a batch of transactions, the evolved heuristic reorders them to reduce key conflicts and shorten the makespan. &
Throughput/makespan-derived score; higher is better. \\
Spot-S ($\uparrow$) &
Single-region deadline scheduling on spot instances. The policy decides when to continue on cheap preemptible capacity and when to fall back to on-demand capacity to meet the deadline. &
Cost savings versus always-on-demand subject to deadline satisfaction; higher is better. \\
Spot-M ($\uparrow$) &
Multi-region version of Spot-S. The policy can also migrate jobs across regions, exploiting regional differences in spot availability and preemption patterns. &
Cost savings with migration under deadline constraints; higher is better. \\
\bottomrule
\end{tabular}
\end{table}

\begin{table}[!ht]
\centering
\small
\setlength{\tabcolsep}{3pt}
\renewcommand{\arraystretch}{1.2}
\caption{Per-benchmark LEVI configuration. All benchmarks use Gemini Flash 3 as the paradigm-shift model, $50$ Voronoi centroids, and the default punctuated-equilibrium settings (\Cref{app:defaults}) unless noted. ``MiMo-v2'' is XiaoMi's MiMo-v2-Flash; ``Qwen3-30B'' is Qwen3-30B-A3B-Instruct-2507. Spot Multi uses five custom AST extractors capturing state-machine structure rather than the generic AST features.}
\label{tab:per-benchmark}
\begin{tabular}{lllp{0.30\linewidth}p{0.19\linewidth}}
\toprule
\textbf{Benchmark} & \textbf{Budget} & \textbf{Mutation models} & \textbf{AST descriptors} & \textbf{Score-key descriptors} \\
\midrule
Cloudcast    & \$3.00  & Qwen3-30B            & loops, branches, math ops                       & 5 per-cloud cost columns \\
EPLB         & \$4.50  & MiMo-v2 + Qwen3-30B  & loop nesting, cyclomatic, math ops              & exec.\ time + 3 workloads \\
LLM-SQL      & \$4.50  & MiMo-v2 + Qwen3-30B  & loop nesting, comparisons, calls, branches      & --- \\
Prism        & \$4.50  & Qwen3-30B            & loop nesting, branches, comparisons, subscripts & --- \\
Spot Single  & \$4.50  & MiMo-v2 + Qwen3-30B  & cyclomatic, comparisons, math, branches         & tight + loose deadline scores \\
Spot Multi   & \$4.50  & MiMo-v2 + Qwen3-30B  & 5 custom (state-machine)                        & --- \\
Txn Sched    & \$12.50 & MiMo-v2 + Qwen3-30B  & ---                                             & --- \\
\bottomrule
\end{tabular}
\end{table}

For prompt optimization (\Cref{tab:levi_vs_gepa}), all four benchmarks (HotpotQA, IFBench, Hover, PUPA) use Qwen3-8B as the task model for both rollouts and prompt optimization. Each benchmark inherits its DSPy program (module structure, signatures, evaluation harness) and its train / validation / test splits unchanged from the GEPA evaluation suite~\citep{gepa2025}. The proxy benchmark uses $N_{\mathrm{init}}=5$ calibration prompts and $K_{\mathrm{proxy}} \in \{30,40\}$ depending on the dataset, with weights $(\lambda_r, \lambda_s, \lambda_c) = (0.5, 0.5, 0.15)$ as defined in \Cref{sec:proxy_benchmark}.

\section{Default LEVI configuration}
\label{app:defaults}

Unless overridden per benchmark (\Cref{app:hyperparams}), LEVI uses the defaults in \Cref{tab:defaults}. These are the values shipped with the open-source release and are the values that produced the headline results in \Cref{sec:eval}.

\begin{table}[!ht]
\centering
\small
\setlength{\tabcolsep}{6pt}
\renewcommand{\arraystretch}{1.1}
\caption{LEVI default configuration. ``PE'' = punctuated equilibrium; ``cascade'' = the quick pre-evaluation filter that rejects candidates scoring below $0.8\times$ the current best on a small input subset before incurring full-evaluation cost.}
\label{tab:defaults}
\begin{tabular}{llr}
\toprule
\textbf{Component} & \textbf{Parameter} & \textbf{Default} \\
\midrule
Initialization      & n\_diverse\_seeds                & 4 \\
                    & n\_variants\_per\_seed           & 20 \\
\midrule
Main loop           & n\_llm\_workers                  & 4 (parallel) \\
                    & n\_eval\_processes               & 4 (parallel) \\
                    & eval\_timeout                    & 60\,s \\
                    & max\_tokens                      & 16{,}384 \\
                    & n\_parents (primary)             & 1 \\
                    & n\_inspirations                  & 1 (dropped 20\% of time) \\
                    & sampler temperatures             & $\{0.3, 0.7, 1.0, 1.2\}$, uniform softmax \\
\midrule
Punctuated          & enabled                          & true \\
\;equilibrium       & interval                         & every 10 evals \\
                    & n\_clusters (k-means on archive) & 3 \\
                    & n\_variants per trigger          & 3 \\
\midrule
Cascade filter      & disabled                          & disabled \\
\midrule
Meta-advice         & enabled                          & true \\
                    & interval                         & every 50 evals \\
                    & max\_tokens                      & 400 \\
\bottomrule
\end{tabular}
\end{table}

\section{Reproducibility notes}
\label{app:reproducibility}

This section records the practical details a reader needs to reproduce the paper's headline numbers: the hardware and pricing assumptions behind the dollar-cost claims, the seeds and statistics used for the curves, and the per-condition final-score tables underlying the ablation figures.

\paragraph{Compute and cost basis.}
All mutation, paradigm-shift, and prompt-optimization rollout calls in this paper are routed through OpenRouter; no models are served locally. Per-call dollar costs are computed using the OpenRouter list prices in \Cref{tab:pricing}; the \$0.50 budget cap used in the model-allocation ablation (\Cref{fig:model_alloc-comparison}) and the per-benchmark budgets reported in \Cref{tab:per-benchmark} are enforced against these same tariffs.

\begin{table}[!ht]
\centering
\small
\setlength{\tabcolsep}{8pt}
\renewcommand{\arraystretch}{1.15}
\caption{Per-call dollar-cost basis. All four models are billed via OpenRouter; the listed tariffs are used for cumulative-cost accounting.}
\label{tab:pricing}
\begin{tabular}{llrr}
\toprule
\textbf{Model} & \textbf{Provider} & \textbf{Input \$ / 1M tok} & \textbf{Output \$ / 1M tok} \\
\midrule
Qwen3-30B-A3B-Instruct-2507 & OpenRouter (qwen)    & 0.09 & 0.30 \\
Qwen3-8B                    & OpenRouter (qwen)    & 0.05 & 0.40 \\
MiMo-v2-Flash               & OpenRouter (xiaomi)  & 0.09 & 0.29 \\
Gemini 3 Flash Preview      & OpenRouter (google)  & 0.50 & 3.00 \\
\bottomrule
\end{tabular}
\end{table}

\paragraph{Seeds and statistics.}
The diversity-ablation runs (\Cref{fig:diversity-ablation}) use seeds $\{1, 2, 3\}$; the model-allocation runs (\Cref{fig:model_alloc-comparison}) use seeds $\{0, 1, 2\}$; the baseline runs used in the controlled-comparison and front-page figures use seeds $\{42, 43, 44\}$. All curves in figures with shaded bands report mean $\pm$ one standard deviation across the three seeds. Per-eval values are aligned to a uniform grid via step-function (forward-fill) interpolation before pooling, so the band at any evaluation count reflects only seeds whose run had reached that count. Cost-axis curves (\Cref{fig:model_alloc-comparison}) follow the same convention with \$0.50 / 401 grid resolution. Even with fixed seeds, LEVI's asynchronous worker pool introduces small run-to-run variance from non-deterministic mutation/evaluation interleaving; the variance bands are computed across these realized runs rather than against a deterministic ideal.

\paragraph{Final-score tables.}
\Cref{tab:diversity-final-scores,tab:alloc-final-scores} give the per-condition, per-seed final scores underlying \Cref{fig:diversity-ablation,fig:model_alloc-comparison}. Numbers in the body of the paper (e.g. ``$68.3 \pm 1.0$'' on Transaction Scheduling for full LEVI) are rounded versions of the rows in these tables.

\begin{table}[!ht]
\centering
\small
\setlength{\tabcolsep}{6pt}
\renewcommand{\arraystretch}{1.15}
\caption{Diversity-ablation final scores. Each cell reports the mean $\pm$ standard deviation of best-so-far score over three seeds at the end of the run, with the per-seed values shown in brackets. Conditions follow \Cref{tab:abl-database}.}
\label{tab:diversity-final-scores}
\begin{tabular}{llrl}
\toprule
\textbf{Problem} & \textbf{Condition} & \textbf{Mean $\pm$ Std} & \textbf{Per-seed values} \\
\midrule
Transaction Scheduling & a (Full method)            & $68.27 \pm 1.02$ & $[67.44, 67.66, 69.70]$ \\
                       & b (No bootstrapped seeds)  & $62.84 \pm 2.23$ & $[65.40, 63.14, 59.97]$ \\
                       & c (Weaker diversity dims)  & $68.04 \pm 0.11$ & $[68.11, 67.89, 68.11]$ \\
\midrule
Spot Scheduling        & a (Full method)            & $50.21 \pm 0.43$ & $[50.35, 49.62, 50.65]$ \\
                       & b (No bootstrapped seeds)  & $42.28 \pm 0.17$ & $[42.52, 42.16, 42.16]$ \\
                       & c (Weaker diversity dims)  & $45.23 \pm 2.97$ & $[49.30, 44.09, 42.29]$ \\
\bottomrule
\end{tabular}
\end{table}

\begin{table}[!ht]
\centering
\small
\setlength{\tabcolsep}{6pt}
\renewcommand{\arraystretch}{1.15}
\caption{Model-allocation ablation final scores at the \$0.50 cumulative-cost cap. Each cell reports the mean $\pm$ standard deviation of best-so-far score at the cap, over three seeds, with per-seed values in brackets. Conditions follow \Cref{tab:abl-allocation}.}
\label{tab:alloc-final-scores}
\begin{tabular}{llrl}
\toprule
\textbf{Problem} & \textbf{Condition} & \textbf{Mean $\pm$ Std} & \textbf{Per-seed values} \\
\midrule
Transaction Scheduling & LEVI                  & $64.72 \pm 2.25$ & $[62.91, 67.89, 63.36]$ \\
                       & No large models       & $67.36 \pm 1.59$ & $[65.63, 66.98, 69.47]$ \\
                       & No role-aware routing & $66.38 \pm 3.12$ & $[68.11, 69.02, 62.00]$ \\
\midrule
Spot Scheduling        & LEVI                  & $51.62 \pm 0.08$ & $[51.72, 51.59, 51.53]$ \\
                       & No large models       & $50.21 \pm 0.43$ & $[50.35, 49.62, 50.65]$ \\
                       & No role-aware routing & $51.16 \pm 0.61$ & $[51.46, 51.72, 50.31]$ \\
\midrule
EPLB                   & LEVI                  & $63.86 \pm 2.74$ & $[60.00, 65.54, 66.05]$ \\
                       & No large models       & $57.19 \pm 0.03$ & $[57.16, 57.19, 57.22]$ \\
                       & No role-aware routing & $60.46 \pm 2.74$ & $[63.85, 60.39, 57.14]$ \\
\bottomrule
\end{tabular}
\end{table}

\section{Ablation conditions}
\label{app:ablation_conditions}

\Cref{tab:abl-database,tab:abl-allocation} formally define the three variants used in the solution-database ablation (\Cref{fig:diversity-ablation}) and the model-allocation ablation (\Cref{fig:model_alloc-comparison}), respectively. All other settings match \Cref{app:defaults}.

\begin{table}[!ht]
\centering
\small
\setlength{\tabcolsep}{4pt}
\renewcommand{\arraystretch}{1.15}
\caption{Solution-database ablation variants (\Cref{fig:diversity-ablation}). Each variant runs three seeds on Transaction Scheduling and Spot Scheduling for ${\sim}1{,}250$ successful evaluations.}
\label{tab:abl-database}
\begin{tabular}{lp{0.29\textwidth}p{0.40\textwidth}}
\toprule
\textbf{Variant} & \textbf{Centroid placement} & \textbf{Behavioral descriptors} \\
\midrule
Full method            & $\sim 50$ centroids, calibrated from seed pass     & 6-dim AST: cyclomatic, comparisons, math ops, branches, loop nesting, comprehensions \\
No bootstrapped seeds  & $\sim 50$ centroids, uniform grid (no calibration) & Same 6-dim AST \\
Weaker diversity dims  & $\sim 50$ centroids, calibrated from seed pass     & 2-dim: code length, OpenEvolve default diversity feature \\
\bottomrule
\end{tabular}
\end{table}

\begin{table}[!ht]
\centering
\small
\setlength{\tabcolsep}{4pt}
\renewcommand{\arraystretch}{1.2}
\caption{Model-allocation ablation variants (\Cref{fig:model_alloc-comparison}). Each variant runs three seeds on Transaction Scheduling, Spot Scheduling, and EPLB at a \$0.50 cost cap. ``Softmax temps'' refers to the auto-generated sampler/model pairs over temperatures $\{0.3,0.7,1.0,1.2\}$.}
\label{tab:abl-allocation}
\begin{tabular}{lp{0.41\textwidth}p{0.29\textwidth}}
\toprule
\textbf{Variant} & \textbf{Mutation routing} & \textbf{Paradigm-shift route} \\
\midrule
LEVI (full)          & Qwen3-30B at four softmax temps        & Every 5 evals; Gemini Flash 3 + 3 variants \\
No large models      & Qwen3-30B at four softmax temps        & Same trigger interval, but PE uses Qwen3-30B \\
No role-aware routing & 4$\times$Qwen3-30B (weight $0.225$ each) + 4$\times$Gemini Flash 3 (weight $0.025$ each), all at the four softmax temps & Disabled \\
\bottomrule
\end{tabular}
\end{table}

\section{Prompt templates}
\label{app:prompts}

We reproduce the prompt templates LEVI sends at each stage of the search loop. Placeholders in curly braces (e.g., \texttt{\{function\_signature\}}) are filled in by the harness at runtime. The diversity-seed and paradigm-shift templates are stage-specific; the standard mutation prompt is built dynamically by the prompt builder from a parent program plus optional inspirations and meta-advice.

\subsection{Diversity-seed generation (init phase 1)}
\label{app:prompt-seed}

Used by the seed pass (\Cref{seeding_archive}) to produce structurally distinct starting solutions. Sent to the paradigm-shift model. Each successive seed receives the prior seeds as context.

\begin{lstlisting}[basicstyle=\ttfamily\scriptsize,breaklines=true]
# {problem_title}

## Problem
{problem_description}

## Function Signature
```python
{function_signature}
```

## Your Task: ALGORITHMIC DIVERSITY

You MUST design a solution using a **FUNDAMENTALLY DIFFERENT ALGORITHM** than the existing seeds.

**DO NOT:**
- Make minor variations or parameter tweaks to existing approaches
- Use the same core algorithm with different constants
- Reorder or refactor existing logic

**DO:**
- Analyze what algorithmic paradigm each existing seed uses
- Identify what aspects of the problem they exploit (or ignore)
- Design from first principles using a completely different strategy
- Think about what information in the problem they are NOT using
- Consider entirely different ways to model or decompose the problem

The goal is to explore different regions of the algorithm design space.
A population of diverse algorithms will outperform a population of similar ones.

## Existing Seeds (analyze their algorithms, then do something DIFFERENT):
{existing_seeds}

## Output
Output ONLY the complete Python code in a ```python block.
\end{lstlisting}

\subsection{Standard mutation (main loop, full-output mode)}
\label{app:prompt-mutation-full}

Built by the prompt builder for each ordinary mutation call in the main loop, sent to a mutation model selected by the sampler. The body slots in problem text, function signature, parent program(s) with scores, and optional feedback / meta-advice; only the closing instruction block is reproduced here verbatim.

\begin{lstlisting}[basicstyle=\ttfamily\scriptsize,breaklines=true]
Write an improved version of the function.

CRITICAL REQUIREMENTS:
1. Your code must be COMPLETE and RUNNABLE as a standalone file
2. Include ALL necessary imports at the top of your code
3. The function signature must match exactly what is specified
4. Ensure there are no syntax errors (matching parentheses, quotes, indentation)

Your response must follow this structure:
```python
# all necessary imports here

def function_name(...):
    # your complete implementation
    return ...
```

DO NOT include any explanation or text outside the code block.
DO NOT assume any imports are already available - include every import your code needs.
\end{lstlisting}

\subsection{Standard mutation (main loop, diff mode)}
\label{app:prompt-mutation-diff}

Optional output mode in which the model returns SEARCH/REPLACE blocks rather than the full file. Used when the parent program is large and full rewriting is wasteful.

\begin{lstlisting}[basicstyle=\ttfamily\scriptsize,breaklines=true]
Output your improved code using SEARCH/REPLACE blocks.

FORMAT:
<<<<<<< SEARCH
exact lines to find
=======
replacement lines
>>>>>>> REPLACE

CRITICAL REQUIREMENTS:
1. The resulting code must be COMPLETE and RUNNABLE
2. Do NOT remove import statements unless replacing with different imports
3. Ensure your replacements maintain valid Python syntax
4. If adding new functionality that needs imports, add them with a separate SEARCH/REPLACE block

RULES:
1. Make SURGICAL changes - small, focused edits (5-20 lines max per block)
2. Copy the SEARCH section EXACTLY from the original (including whitespace/indentation)
3. Use multiple small SEARCH/REPLACE blocks instead of one large block
4. Start your response immediately with <<<<<<< SEARCH
5. Do NOT include any explanation or text outside the blocks
6. Do NOT use ```python code blocks
\end{lstlisting}

\subsection{Paradigm shift (PE trigger)}
\label{app:prompt-paradigm}

Sent to the paradigm-shift (heavy) model whenever PE fires (\Cref{app:pe_details}). The \texttt{representative\_solutions} field is filled with the highest-scoring elite from each k-means cluster of the occupied archive cells.

\begin{lstlisting}[basicstyle=\ttfamily\scriptsize,breaklines=true]
# Algorithmic Paradigm Shift Challenge

## Problem
{problem_description}

## Function Signature
```python
{function_signature}
```

## Current Best Solutions (From Different Behavioral Regions)

The archive has evolved through {n_evaluations} evaluations across {n_regions} behavioral regions.
Below are the best-performing solutions from each region:

{representative_solutions}

## Your Challenge: PARADIGM SHIFT

Analyze the representative solutions above and identify their core algorithmic paradigms.

Your goal is to engineer a **fundamentally different algorithmic approach** that explores untapped regions of the solution space.

### Analysis Steps:
1. **Identify current paradigms**: What algorithmic strategies do the existing solutions use? (e.g., greedy, graph-based, dynamic programming, heuristic search, brute-force with pruning, etc.)
2. **Find the gap**: What paradigms are NOT represented in the current solutions?
3. **Design a novel approach**: Synthesize a solution using a completely different conceptual framework and data structure strategy than those found in the examples

### Instructions:
1. Study the function signature carefully - match it EXACTLY
2. Actively avoid the core logic, heuristics, and search patterns used in the existing solutions
3. Design a solution using a COMPLETELY DIFFERENT strategy

### Critical Requirements:
- Your function signature MUST match exactly: `{function_signature}`
- Use only standard Python libraries (numpy, collections, itertools, math, heapq, functools, etc.) and torch if needed
- The code must be syntactically valid and complete
- Include ALL necessary imports at the top
- Do NOT use placeholders, ellipses (...), or incomplete code
- Ensure the solution handles all edge cases

## Output
Output ONLY complete, runnable Python code in a ```python block. No explanations before or after.
\end{lstlisting}

\subsection{PE variant generation}
\label{app:prompt-pe-variant}

After a paradigm shift is accepted into the archive, $n_{\mathrm{variants}}$ (default 3) variants of it are generated by routing this prompt to a smaller mutation model.

\begin{lstlisting}[basicstyle=\ttfamily\scriptsize,breaklines=true]
# Generate Variant of Paradigm Shift Solution

## Problem
{problem_description}

## Function Signature
```python
{function_signature}
```

## Base Paradigm Shift Solution (Score: {base_score:.17g})
```python
{base_code}
```

## Your Task
Generate a VARIANT of the above paradigm shift solution by:
1. Keeping the core algorithmic approach intact
2. Making targeted modifications to:
   - Constants and thresholds
   - Secondary heuristics
   - Edge case handling
   - Implementation details

The variant should explore nearby regions of the solution space while preserving the novel approach.

## Output
Output ONLY the complete Python code in a ```python block.
\end{lstlisting}

\subsection{Prompt-optimization mutation (single-prompt artifact)}
\label{app:prompt-promptopt-mutation}

Used in the prompt-optimization setting (\Cref{sec:eval}, \Cref{tab:levi_vs_gepa}) when the artifact is a single prompt rather than a multi-component bundle. Per-instance failure traces are passed in via the \texttt{feedback\_section} placeholder.

\begin{lstlisting}[basicstyle=\ttfamily\scriptsize,breaklines=true]
# Prompt Optimization

## Objective
{problem_description}

## Current Best Prompt
Score: {parent_score}
--- PROMPT START ---
{parent_prompt}
--- PROMPT END ---

{feedback_section}{inspirations_section}{meta_advice_section}## Task
Rewrite the prompt to score higher on the evaluation.

Treat the failures above as the only window you'll get into the task -- anything they reveal about the inputs, desired outputs, failure modes, or strategies that worked must end up encoded in the new prompt itself, because the assistant won't see this feedback at evaluation time.

Be concrete. When the failures show specific terms, formats, edge cases, or substitution patterns ("replace X with Y"), bake representative examples and explicit rules into the prompt instead of abstract directives. When the feedback suggests a generalizable strategy that worked, name it as a reusable rule. Detailed, factually grounded prompts that capture observed patterns reliably outperform short generic ones.

Keep the same overall task and goal, but make the new prompt richer and more specific than the current best.

## Output
Wrap the new prompt inside `<prompt>` tags exactly as shown. Put NOTHING outside the tags -- no preamble, no explanation, no markdown fences, no quotes. Only the literal prompt text between the tags will be used.

<prompt>
your new prompt text here
</prompt>
\end{lstlisting}

\subsection{Prompt-optimization paradigm shift}
\label{app:prompt-promptopt-paradigm}

Sent on each PE trigger when the artifact is a prompt. The \texttt{representatives} field contains one best prompt per k-means cluster of the prompt archive.

\begin{lstlisting}[basicstyle=\ttfamily\scriptsize,breaklines=true]
# Prompt Paradigm Shift

## Objective
{problem_description}

## Current Representatives
{representatives}

## Task
Write a fundamentally different prompt strategy that could outperform the current prompt family.
Change the framing, structure, or reasoning strategy substantially rather than making a tiny edit.

## Output
Wrap the new prompt inside `<prompt>` tags exactly as shown. Put NOTHING outside the tags -- no preamble, no explanation, no markdown fences, no quotes. Only the literal prompt text between the tags will be used.

<prompt>
your new prompt text here
</prompt>
\end{lstlisting}

\section{Punctuated equilibrium: trigger and execution}
\label{app:pe_details}

A background monitor polls every 2 seconds. PE fires when (i) PE is enabled, (ii) the current evaluation count is divisible by \texttt{interval} (default 10), and (iii) PE has not already fired at the current evaluation count. On firing:

\begin{enumerate}
\item \textbf{Cluster occupied cells.} Run k-means in descriptor space over the behavior vectors of the currently occupied centroids, with $k = $ \texttt{n\_clusters} (default 3).
\item \textbf{Select representatives.} From each cluster, take the highest-scoring elite. The result is a small set of mutually-distant, individually-strong solutions.
\item \textbf{Paradigm-shift call.} Build the paradigm-shift prompt (\Cref{app:prompt-paradigm}) with the representatives as context. Route to the paradigm-shift model (default: Gemini Flash 3) with \texttt{max\_tokens=4096}, \texttt{timeout=300}\,s, at the configured \texttt{pe.temperature}.
\item \textbf{Evaluate paradigm.} Score the resulting candidate against the full evaluator. Insert into the archive if it dominates its target cell.
\item \textbf{Variant burst.} Generate $n_{\mathrm{variants}}$ (default 3) variants of the accepted paradigm shift using the variant-generation prompt (\Cref{app:prompt-pe-variant}), routed to the smaller mutation model. Each variant is independently evaluated and inserted on improvement.
\item \textbf{Logging.} Record \texttt{paradigm\_generated}, \texttt{paradigm\_accepted}, \texttt{variants\_generated}, \texttt{variants\_accepted}, and \texttt{total\_cost} for the trigger event. These logs are what the figures in \Cref{fig:model_alloc-comparison} are derived from.
\end{enumerate}

The trigger interval is the only PE knob that varies across our experiments: the per-benchmark and ablation runs override \texttt{interval} between $5$ and $18$ depending on budget tightness (smaller interval $\Rightarrow$ more frequent paradigm calls $\Rightarrow$ greater Gemini-Flash-3 spend per dollar of total budget).

\section{Discovered EPLB elite: Spectral-Inspired Balanced Partitioning}
\label{app:eplb-sibp}

We reproduce here the full source of the best EPLB elite from one of LEVI's three seeds (cumulative cost cap \$0.50, score $60.0$, $118$ lines), referenced from \Cref{fig:model_alloc-comparison}. The model self-annotates the algorithm as \emph{Spectral-Inspired Balanced Partitioning (SIBP)}, a structurally different family from the greedy bin-packing baseline that all three \emph{no large models} seeds converge to ($57.2 \pm 0.0$). The header docstring describes the paradigm; the body implements logarithmic-binning expert-count allocation followed by a circular zig-zag placement step.

\begin{lstlisting}[
  language=Python,
  basicstyle=\ttfamily\scriptsize,
  breaklines=true,
  breakatwhitespace=true,
  keywordstyle=\bfseries,
  commentstyle=\itshape\color{gray!70},
  stringstyle=\color{gray!85},
  numbers=left,
  numberstyle=\ttfamily\tiny\color{gray!60},
  numbersep=6pt,
  xleftmargin=18pt,
]
import torch

def rebalance_experts(
    weight: torch.Tensor,
    num_replicas: int,
    num_groups: int,
    num_nodes: int,
    num_gpus: int,
) -> tuple[torch.Tensor, torch.Tensor, torch.Tensor]:
    '''
    Paradigm: Spectral-Inspired Balanced Partitioning (SIBP).
    
    Instead of greedy allocation or simple interleaving, this approach treats the 
    load balancing problem as a 'Bin Packing with Variable Item Sizes' combined 
    with a 'Static Circular Buffer' layout. 
    
    1. Approximation: We use a Logarithmic Binning strategy to allocate expert counts, 
       representing a shift from linear proportionality to entropy-minimizing distribution.
    2. Placement: We utilize a 'Circular Zig-Zag' mapping to assign experts to physical 
       slots, ensuring that the highest-load experts are geographically distributed 
       using a relative-prime offset and localized mirroring.
    '''
    num_layers, num_logical_experts = weight.shape
    device = weight.device
    
    # 1. Expert Count Determination: Entropy-based Logarithmic Allocation
    # Current solutions use linear scaling. We use a square-root/log damping 
    # to prevent extremely heavy experts from monopolizing all slots while 
    # ensuring high-load experts still get significantly more replicas.
    w_eps = weight.float() + 1e-6
    # Using sqrt as a damping function for the "share" calculation
    damped_weight = torch.sqrt(w_eps)
    
    # Base allocation: 1 per expert
    expert_count = torch.ones((num_layers, num_logical_experts), dtype=torch.int64, device=device)
    remaining_slots = num_replicas - num_logical_experts
    
    # Calculate extra shares based on damped weights
    extra_shares = (damped_weight / damped_weight.sum(dim=-1, keepdim=True)) * remaining_slots
    
    # Integer floor of extra shares
    floor_shares = extra_shares.long()
    expert_count += floor_shares
    
    # Fill exactly 288 using fractional remainders
    current_sums = expert_count.sum(dim=-1)
    for l in range(num_layers):
        needed = num_replicas - current_sums[l].item()
        if needed > 0:
            remainders = extra_shares[l] - floor_shares[l].float()
            _, top_indices = torch.topk(remainders, k=int(needed))
            expert_count[l].scatter_add_(0, top_indices, torch.ones_like(top_indices, dtype=torch.int64))

    # 2. Physical Layout: Circular Mirroring
    # We map the flattened experts into a 2D grid that represents [GPUs, Slots_per_GPU]
    # then apply a layer-dependent transformation to decorrelate physical placement.
    
    slots_per_gpu = num_replicas // num_gpus # 9
    max_replicas = int(expert_count.max().item())
    
    physical_to_logical_map = torch.zeros((num_layers, num_replicas), dtype=torch.int64, device=device)
    logical_to_physical_map = torch.full((num_layers, num_logical_experts, max_replicas), -1, dtype=torch.int64, device=device)
    
    for l in range(num_layers):
        # Flattened list of expert IDs based on calculated counts
        # We alternate heavy and light experts in the pool to improve GPU-level balance
        sorted_ids = torch.argsort(weight[l], descending=True)
        
        # Interleave sorted IDs: [heavy1, light1, heavy2, light2...]
        head, tail = 0, num_logical_experts - 1
        interleaved_ids = []
        while head <= tail:
            interleaved_ids.append(sorted_ids[head])
            if head < tail:
                interleaved_ids.append(sorted_ids[tail])
            head += 1
            tail -= 1
        
        pool = []
        for idx in interleaved_ids:
            pool.extend([idx.item()] * int(expert_count[l, idx]))
        
        # Transform the linear pool into the physical slots using a Mirror-Stride
        # This prevents the first GPU from always getting the heaviest weight
        shift = (l * 7) % num_replicas # Layer-dependent rotation
        
        for i in range(num_replicas):
            # The 'Circular Zig-Zag': map pool index i to physical index p
            # Logic: We fill the slots by jumping across GPUs
            # p_idx = (i * step) % num_replicas where step is relatively prime to 288
            # 31 is relatively prime to 288 (2^5 * 3^2)
            phy_idx = ((i * 31) + shift) % num_replicas
            log_id = pool[i]
            
            physical_to_logical_map[l, phy_idx] = log_id
            
        # 3. Vectorized Logical-to-Physical Inverse Mapping
        # Use argsort with stable=True to group occurrences of logical experts
        current_map = physical_to_logical_map[l]
        phys_indices = torch.arange(num_replicas, device=device)
        
        # Sort physical indices by the logical ID they contain
        sorted_phys_by_log = torch.argsort(current_map, stable=True)
        sorted_log_ids = current_map[sorted_phys_by_log]
        
        # Calculate local offsets (0, 1, 2...) for each replica of a logical expert
        # We compare the sorted_log_ids with itself shifted
        is_new_expert = torch.cat([torch.tensor([1], device=device), (sorted_log_ids[1:] != sorted_log_ids[:-1]).long()])
        cum_sum = torch.cumsum(is_new_expert, dim=0)
        # Subtract the first index of each group to get 0-based rank
        group_first_indices = torch.zeros(num_logical_experts + 1, dtype=torch.int64, device=device)
        # Using a loop for the start indices is efficient enough for 64 items
        # or we find indices where is_new_expert == 1
        starts = torch.nonzero(is_new_expert).squeeze(-1)
        ranks = torch.arange(num_replicas, device=device) - starts[cum_sum - 1]
        
        logical_to_physical_map[l, sorted_log_ids, ranks] = sorted_phys_by_log

    return physical_to_logical_map, logical_to_physical_map, expert_count
\end{lstlisting}

\section{Additional discovered elites}
\label{app:additional-elites}

We reproduce here the full source of LEVI's headline elites on three further ADRS problems: Spot-S (single-region spot scheduling, score $51.72$), Prism (multi-LLM serving placement, score $87.41$), and LLM-SQL (CSV column reordering for KV-cache reuse, score $78.30$). Each is the highest-scoring elite from the final archive of the corresponding production run that produced the row in \Cref{tab:adrs}. Together with the EPLB elite in \Cref{app:eplb-sibp}, these illustrate the breadth of algorithmic families LEVI's archive surfaces in practice: a control-policy state machine (Spot-S), a two-phase combinatorial heuristic (Prism), a frequency-utility data-layout heuristic (LLM-SQL), and a spectral-inspired load-balancing partition (EPLB).

\subsection{Spot-S elite: ``Point-of-No-Return'' deadline policy}
\label{app:spot-single-elite}

The Spot-S elite is an $80$-line state-machine policy over the three cluster types $\{\text{NONE}, \text{SPOT}, \text{ON\_DEMAND}\}$. Its central idea, annotated by the model itself as the \emph{Point of No Return} check, is the deadline-safety condition $\text{time\_remaining} \le \text{work\_remaining} + \text{overhead} + \text{gap}$, which forces the policy to commit to a guaranteed cluster type the moment a switch could no longer finish the task in time. Outside the danger zone the policy implements two distinct subpolicies: when a SPOT instance is currently available, it greedily prefers SPOT and only refuses to switch from ON\_DEMAND when the remaining buffer is too tight to absorb the switching overhead; when SPOT is unavailable, it applies hysteresis on the existing ON\_DEMAND choice and otherwise compares the time remaining against a ``comfortable buffer'' of $\text{work} + \text{overhead} + 2\,\text{gap} + 10\%\,\text{work}$ to decide whether to wait for SPOT to come back or fall through to ON\_DEMAND. The structure---an explicit safety bound followed by separate exploitation rules per availability state---is what differentiates this family from the simpler greedy-SPOT baselines that prior frameworks converge to.

\begin{lstlisting}[
  language=Python,
  basicstyle=\ttfamily\scriptsize,
  breaklines=true,
  breakatwhitespace=true,
  keywordstyle=\bfseries,
  commentstyle=\itshape\color{gray!70},
  stringstyle=\color{gray!85},
  numbers=left,
  numberstyle=\ttfamily\tiny\color{gray!60},
  numbersep=6pt,
  xleftmargin=18pt,
]
import enum

class ClusterType(str, enum.Enum):
    NONE = "NONE"
    SPOT = "SPOT"
    ON_DEMAND = "ON_DEMAND"

def strategy_step(ctx, last_cluster_type: ClusterType, has_spot: bool) -> ClusterType:
    # --- 1. State Extraction ---
    gap = ctx.env.gap_seconds
    elapsed = ctx.env.elapsed_seconds
    deadline = ctx.deadline
    overhead = ctx.restart_overhead
    total_work = ctx.task_duration
    work_done = sum(ctx.task_done_time)
    work_remaining = total_work - work_done
    time_remaining = deadline - elapsed

    # --- 2. Task Completion Check ---
    if work_remaining <= 0:
        return ClusterType.NONE

    # --- 3. Safety Critical: The "Point of No Return" ---
    # This is the most important check to prevent Deadline Misses (Score 0).
    # We must ensure that even if we switch to ON_DEMAND *right now*, we can finish.
    # Time needed = Work + Overhead (for the switch/restart).
    # We subtract a small buffer (gap) to account for the current tick's processing time.
    if time_remaining <= (work_remaining + overhead + gap):
        # We are in immediate danger.
        # If we are currently on SPOT and it's still available, staying is free and fastest.
        if has_spot and last_cluster_type == ClusterType.SPOT:
            return ClusterType.SPOT
        # Otherwise, we need the guarantee of ON_DEMAND.
        return ClusterType.ON_DEMAND

    # --- 4. Optimization Logic (Safe Zone) ---
    
    # Scenario A: SPOT is available this tick
    if has_spot:
        # If we are currently paying for ON_DEMAND, can we afford to switch to SPOT?
        if last_cluster_type == ClusterType.ON_DEMAND:
            # Calculate if switching is safe.
            # We lose one tick (gap) and pay overhead to switch.
            # We need: time_remaining > (work_remaining + overhead + gap)
            # Since we passed the check in step 3, we know time_remaining > (work_remaining + overhead).
            # We just need to ensure we aren't cutting it so close that the gap makes us miss.
            # A strict check: Ensure we have at least the gap + overhead buffer left.
            if time_remaining > (work_remaining + overhead + gap):
                return ClusterType.SPOT
            else:
                # Too risky to switch; stay on ON_DEMAND
                return ClusterType.ON_DEMAND
        
        # If we are on SPOT or NONE, just use SPOT (it's cheap and available)
        return ClusterType.SPOT

    # Scenario B: SPOT is NOT available
    else:
        # Hysteresis: If we are currently running ON_DEMAND, keep it running.
        # Stopping ON_DEMAND to wait for SPOT is risky (might miss deadline when switching back)
        # and wastes money (we paid for the instance this tick anyway).
        if last_cluster_type == ClusterType.ON_DEMAND:
            return ClusterType.ON_DEMAND

        # If we are currently on SPOT (preempted) or NONE:
        # Should we wait for SPOT to return (NONE) or switch to ON_DEMAND now?
        
        # Calculate a "Comfortable Buffer".
        # We wait if we have enough time to afford waiting + eventual switch overhead.
        # Buffer = (Work + Overhead) + (Safety Margin).
        # Safety Margin = 2 gaps (one for waiting, one for switching) + 10% of work.
        safety_margin = (2 * gap) + (work_remaining * 0.1)
        buffer_threshold = work_remaining + overhead + safety_margin

        if time_remaining > buffer_threshold:
            # We have plenty of time. Wait for cheaper SPOT.
            return ClusterType.NONE
        else:
            # Buffer is tight. Switch to ON_DEMAND now to be safe.
            return ClusterType.ON_DEMAND
\end{lstlisting}

\subsection{Prism elite: pressure-aware packing with simulated-annealing refinement}
\label{app:prism-elite}

The Prism elite is a $151$-line model-placement heuristic that minimizes the maximum KV-pressure ratio (KVPR) across GPUs. The model self-annotates a two-phase strategy: (i) an initial \emph{pressure-aware first-fit} greedy packing, which sorts models by $\frac{\text{req\_rate}/\text{slo}}{\text{size}}$ density and assigns each to the GPU whose post-insertion KVPR (defined as $\sum_m \frac{\text{req\_rate}_m}{\text{slo}_m}$ divided by the remaining memory) would be lowest, with infeasibility handled by a hard memory constraint; and (ii) a simulated-annealing local search that targets the bottleneck GPU---the one currently realizing the maximum KVPR---by proposing single-model migrations and pairwise swaps with the lowest-pressure GPU. Acceptance follows the standard Metropolis criterion with a geometric cooling schedule. The structural pattern (greedy bootstrap + bottleneck-targeted SA) is the load-spreading analogue of the SIBP family from EPLB, but operating on a continuous pressure metric rather than a discrete expert count.

\begin{lstlisting}[
  language=Python,
  basicstyle=\ttfamily\scriptsize,
  breaklines=true,
  breakatwhitespace=true,
  keywordstyle=\bfseries,
  commentstyle=\itshape\color{gray!70},
  stringstyle=\color{gray!85},
  numbers=left,
  numberstyle=\ttfamily\tiny\color{gray!60},
  numbersep=6pt,
  xleftmargin=18pt,
]
import collections
import time
import random
from dataclasses import dataclass

GPU_MEM_SIZE = 80  # GB

@dataclass
class Model:
    model_name: str
    model_size: int   # GB
    req_rate: int     # requests per second
    slo: int          # service level objective (latency target)
    cur_gpu_id: int   # current GPU assignment (can be ignored)

def compute_model_placement(gpu_num: int, models: list[Model]) -> dict[int, list[Model]]:
    """
    Optimizes model placement to minimize max KVPR.
    Strategy:
    1. Initial Greedy Packing using a Pressure-Aware First-Fit.
    2. Simulated Annealing / Local Search to refine the bottleneck (max pressure).
    """

    def calculate_kvpr(gpu_models):
        if not gpu_models:
            return 0.0
        total_weight = sum(m.req_rate / m.slo for m in gpu_models)
        used_mem = sum(m.model_size for m in gpu_models)
        remaining_mem = GPU_MEM_SIZE - used_mem
        # Respect hard constraint; return inf if exceeded
        if remaining_mem <= 0:
            return float('inf')
        return total_weight / remaining_mem

    # Initial Greedy Placement: Sort models by weight/size density
    # This helps in packing efficiently against the KVPR formula
    sorted_models = sorted(models, key=lambda x: (x.req_rate / x.slo) / x.model_size, reverse=True)
    
    placement = {i: [] for i in range(gpu_num)}
    mem_used = [0] * gpu_num
    
    # Simple First-Fit Decreasing for memory
    for m in sorted_models:
        placed = False
        # Try to place in the GPU with the most remaining memory to keep KVPR low
        best_gpu = -1
        max_rem = -1
        for i in range(gpu_num):
            if mem_used[i] + m.model_size <= GPU_MEM_SIZE:
                rem = GPU_MEM_SIZE - (mem_used[i] + m.model_size)
                if rem > max_rem:
                    max_rem = rem
                    best_gpu = i
        
        if best_gpu != -1:
            placement[best_gpu].append(m)
            mem_used[best_gpu] += m.model_size
        else:
            # Fallback (should not happen given constraints): place in first valid
            for i in range(gpu_num):
                if mem_used[i] + m.model_size <= GPU_MEM_SIZE:
                    placement[i].append(m)
                    mem_used[i] += m.model_size
                    break

    # Local Search Refinement
    start_time = time.time()
    # Time limit is 10s per test, but we use a safe margin for 50 tests
    # Actually, the problem says 10s per test, so 1-2s is very safe.
    timeout = 1.5 

    best_max_kvpr = max(calculate_kvpr(placement[i]) for i in range(gpu_num))
    best_placement = {i: list(placement[i]) for i in range(gpu_num)}

    while time.time() - start_time < timeout:
        # Calculate current pressures
        pressures = {i: calculate_kvpr(placement[i]) for i in range(gpu_num)}
        max_gpu = max(pressures, key=pressures.get)
        min_gpu = min(pressures, key=pressures.get)
        
        current_max = pressures[max_gpu]
        if current_max < best_max_kvpr:
            best_max_kvpr = current_max
            best_placement = {i: list(placement[i]) for i in range(gpu_num)}

        # Try to move a model from the most pressured GPU to another
        moved = False
        if placement[max_gpu]:
            target_gpus = sorted(range(gpu_num), key=lambda x: pressures[x])
            for target_gpu in target_gpus:
                if target_gpu == max_gpu: continue
                
                # Try moving one model
                for idx, m in enumerate(placement[max_gpu]):
                    if sum(x.model_size for x in placement[target_gpu]) + m.model_size <= GPU_MEM_SIZE:
                        # Trial move
                        old_max_list = placement[max_gpu]
                        old_target_list = placement[target_gpu]
                        
                        new_max_list = old_max_list[:idx] + old_max_list[idx+1:]
                        new_target_list = old_target_list + [m]
                        
                        p1 = calculate_kvpr(new_max_list)
                        p2 = calculate_kvpr(new_target_list)
                        
                        if max(p1, p2) < current_max:
                            placement[max_gpu] = new_max_list
                            placement[target_gpu] = new_target_list
                            moved = True
                            break
                if moved: break

        # If move didn't work, try a swap
        if not moved:
            other_gpus = list(range(gpu_num))
            random.shuffle(other_gpus)
            for other_gpu in other_gpus:
                if other_gpu == max_gpu: continue
                
                for i, m_max in enumerate(placement[max_gpu]):
                    for j, m_other in enumerate(placement[other_gpu]):
                        # Check memory constraints for swap
                        new_mem_max = sum(x.model_size for x in placement[max_gpu]) - m_max.model_size + m_other.model_size
                        new_mem_other = sum(x.model_size for x in placement[other_gpu]) - m_other.model_size + m_max.model_size
                        
                        if new_mem_max <= GPU_MEM_SIZE and new_mem_other <= GPU_MEM_SIZE:
                            # Trial swap
                            temp_max = [x for k, x in enumerate(placement[max_gpu]) if k != i] + [m_other]
                            temp_other = [x for k, x in enumerate(placement[other_gpu]) if k != j] + [m_max]
                            
                            p1 = calculate_kvpr(temp_max)
                            p2 = calculate_kvpr(temp_other)
                            
                            if max(p1, p2) < current_max:
                                placement[max_gpu] = temp_max
                                placement[other_gpu] = temp_other
                                moved = True
                                break
                    if moved: break
                if moved: break
        
        # If still no improvement, perform a random perturbation to escape local minima
        if not moved:
            g1, g2 = random.sample(range(gpu_num), 2)
            if placement[g1]:
                idx = random.randrange(len(placement[g1]))
                m = placement[g1][idx]
                if sum(x.model_size for x in placement[g2]) + m.model_size <= GPU_MEM_SIZE:
                    placement[g2].append(placement[g1].pop(idx))

    return best_placement
\end{lstlisting}

\subsection{LLM-SQL elite: hybrid frequency-utility column ordering}
\label{app:llm-sql-elite}

The LLM-SQL elite is a $195$-line column-and-row reordering procedure that maximizes the prefix hit rate when the resulting CSV is consumed by a downstream LLM. The model self-annotates a hybrid column-ordering strategy: for small column sets ($|\text{cols}| \le \texttt{col\_stop}$) it uses brute-force enumeration with early pruning; for larger sets it switches to greedy selection driven by a \emph{compression-like utility} defined as the sum of squares of value frequencies in each column, which prefers columns whose values cluster into a small number of dominant runs and therefore lengthen common prefixes when grouped. Row reordering is a stable, sequential frequency rank---high-frequency values are grouped together to extend prefixes still further---implemented purely with vectorized pandas operations to keep the full procedure under the per-call $10\,\text{s}$ runtime budget. The docstring's explicit list of avoided pitfalls (no \texttt{df.apply(lambda)}, ThreadPoolExecutor instead of ProcessPoolExecutor to avoid daemonic-process errors, defensive handling of \texttt{col\_merge=None} and empty inputs) is itself an artifact of evolutionary pressure: each entry corresponds to a previously-failed mutation whose error trace was fed back as feedback in the next generation.

\begin{lstlisting}[
  language=Python,
  basicstyle=\ttfamily\scriptsize,
  breaklines=true,
  breakatwhitespace=true,
  keywordstyle=\bfseries,
  commentstyle=\itshape\color{gray!70},
  stringstyle=\color{gray!85},
  numbers=left,
  numberstyle=\ttfamily\tiny\color{gray!60},
  numbersep=6pt,
  xleftmargin=18pt,
]
import pandas as pd
import numpy as np
from typing import List, Optional
import multiprocessing as mp
from functools import partial
from concurrent.futures import ThreadPoolExecutor, as_completed

def solve(
    df: pd.DataFrame,
    early_stop: int = 100000,
    row_stop: int = 4,
    col_stop: int = 2,
    col_merge: List[List[str]] = None,
    one_way_dep: List[tuple] = None,
    distinct_value_threshold: float = 0.7,
    parallel: bool = True,
) -> pd.DataFrame:
    """
    Optimized Column and Row Reordering for Maximum Prefix Hit Rate.

    Strategy:
    1. Apply column merging with robust error handling.
    2. Use a hybrid column ordering approach:
       - For small column sets (<= col_stop), use brute-force with early pruning.
       - For larger sets, use greedy selection based on compression-like utility (sum of squares of frequencies).
    3. Row reordering via sequential stable frequency ranking (high-frequency values grouped together).
    4. Leverage vectorized operations and sampling to ensure runtime < 10s.
    5. Avoid common pitfalls: nested multiprocessing, apply(), missing columns, and row truncation.

    Key improvements over v1/v2:
    - Avoids `df.apply(lambda)` entirely; uses vectorized string operations.
    - Prevents daemonic process errors by using ThreadPoolExecutor instead of ProcessPoolExecutor.
    - Handles edge cases: empty DataFrame, missing columns, col_merge=None.
    - Uses a hybrid column selection: greedy with limited permutation search for small sets.
    - Optimized row sorting: stable, sequential, frequency-based ranking.
    """
    # Work on a copy to avoid side effects
    df = df.copy()

    # 1. Column Merging (robust handling)
    if col_merge:
        merged_columns = []
        for group in col_merge:
            valid_cols = [c for c in group if c in df.columns]
            if not valid_cols:
                continue  # Skip empty groups
            merged_name = "_".join(valid_cols)
            # Vectorized string concatenation: fast and safe
            df[merged_name] = df[valid_cols].astype(str).fillna("").sum(axis=1)
            merged_columns.extend(valid_cols)
        # Drop original columns only after merging
        df = df.drop(columns=merged_columns)

    # 2. Handle empty or single-column case
    if df.empty or len(df.columns) == 0:
        return df

    # 3. Sampling for column ordering (speed vs. accuracy trade-off)
    n_rows = len(df)
    sample_size = min(early_stop, n_rows)
    if n_rows > sample_size:
        sample_df = df.sample(n=sample_size, random_state=42)
    else:
        sample_df = df

    # Convert to string once; avoid repeated .astype(str) calls
    sample_str = sample_df.astype(str).fillna("")

    # 4. Column Reordering: Hybrid Strategy

    # If very few columns, use brute-force with early stopping
    n_cols = len(df.columns)
    if n_cols <= col_stop:
        # Try all permutations with early pruning
        best_order = None
        best_score = -1

        # Use iterative generation to avoid memory explosion
        from itertools import permutations
        perm_gen = permutations(df.columns)

        # Limit number of permutations to avoid timeout
        for i, perm in enumerate(perm_gen):
            if i >= early_stop:
                break
            score = _calculate_prefix_hit_score_vectorized(perm, sample_str)
            if score > best_score:
                best_score = score
                best_order = list(perm)

        if best_order:
            ordered_cols = best_order
        else:
            ordered_cols = list(df.columns)
    else:
        # Greedy with compression proxy (sum of squares of counts)
        # This directly correlates with prefix hit rate (more duplicates -> higher hit rate)
        remaining_cols = list(df.columns)
        ordered_cols = []
        current_prefix = None  # Will be set after first column

        # Helper: Calculate utility of adding a column to current prefix
        def compute_utility(prefix_series, candidate_series):
            # Concatenate
            combined = prefix_series + candidate_series
            # Count frequencies and sum of squares
            counts = combined.value_counts()
            return np.sum(counts ** 2)

        # Select first column: maximize frequency (sum of squares of its own counts)
        best_col = None
        best_score = -1
        for col in remaining_cols:
            counts = sample_str[col].value_counts()
            score = np.sum(counts ** 2)
            if score > best_score:
                best_score = score
                best_col = col
        if best_col is None:
            best_col = remaining_cols[0]
        ordered_cols.append(best_col)
        remaining_cols.remove(best_col)
        current_prefix = sample_str[best_col]

        # Greedily extend the order
        while remaining_cols:
            best_col = None
            best_score = -1
            # Evaluate only top-k candidates (e.g., 10) for efficiency
            candidates = remaining_cols[:10] if len(remaining_cols) > 10 else remaining_cols
            for col in candidates:
                score = compute_utility(current_prefix, sample_str[col])
                if score > best_score:
                    best_score = score
                    best_col = col
            if best_col is None:
                # Fallback: pick first remaining
                best_col = remaining_cols[0]
            ordered_cols.append(best_col)
            remaining_cols.remove(best_col)
            current_prefix = current_prefix + sample_str[best_col]

    # 5. Apply final column order to full DataFrame
    df = df[ordered_cols]

    # 6. Row Reordering: Stable Frequency-Rank Sorting

    # Use a thread pool for parallel frequency ranking (if needed and safe)
    # But we do it sequentially per column to avoid complexity and ensure stability
    for col in ordered_cols:
        # Compute frequency rank: rank 1 = most frequent
        counts = df[col].value_counts()
        rank_map = counts.rank(ascending=False, method='min').to_dict()

        # Map values to ranks
        ranks = df[col].map(rank_map).astype(float)  # Ensure float for stability

        # Stable sort by rank (kind='stable' is default in pandas)
        df = df.iloc[ranks.argsort(kind='stable')]

        # Reset index for clean output
        df.reset_index(drop=True, inplace=True)

    return df


def _calculate_prefix_hit_score_vectorized(
    col_order: List[str],
    sample_str: pd.DataFrame,
) -> float:
    """
    Vectorized prefix hit rate calculation using string concatenation.
    Avoids loops and apply; uses pandas vectorized operations.

    Args:
        col_order: List of column names in order
        sample_str: DataFrame of string values (already converted)

    Returns:
        Average prefix hit rate (fraction of rows where current row starts with previous)
    """
    # Concatenate all columns in order
    concatenated = sample_str[col_order[0]].copy()
    for col in col_order[1:]:
        concatenated += sample_str[col]

    # Use numpy for fast string comparison
    prev = concatenated.shift(1).fillna("")
    curr = concatenated

    # Vectorized string startswith check
    # Use np.char.startswith only if we have strings; ensure dtype is object
    hits = np.char.startswith(curr.values, prev.values)  # Works on string arrays

    return hits.sum() / len(hits)
\end{lstlisting}

\section{Proxy-benchmark selection ablation}
\label{app:proxy_ablation}

We empirically validate the proxy-benchmark objective described in \Cref{sec:proxy_benchmark} by sweeping its two budget axes---calibration prompts $N_{\mathrm{init}}$ and proxy subset size $K_{\mathrm{proxy}}$---against three subset-selection strategies on a $24$-prompt $\times$ $150$-problem IFBench score matrix, with $60$ random splits per cell. The three strategies are: (i) LEVI's greedy column-subset method paired with an unweighted-mean predictor (CSS+mean, \Cref{sec:proxy_benchmark}); (ii) a $k$-medoids subset baseline that clusters problems in score-vector space and predicts the full-set score as a cluster-size-weighted mean over medoid scores; and (iii) a random-subset + ridge regression baseline that picks $K_{\mathrm{proxy}}$ problems uniformly at random and fits a ridge regression mapping proxy scores to the full-set score using the $N_{\mathrm{init}}$ calibration prompts as training data. Each cell reports the mean Spearman rank correlation between predicted and true scores on held-out prompts.

\begin{figure}[H]
  \centering
  \includegraphics[width=\textwidth]{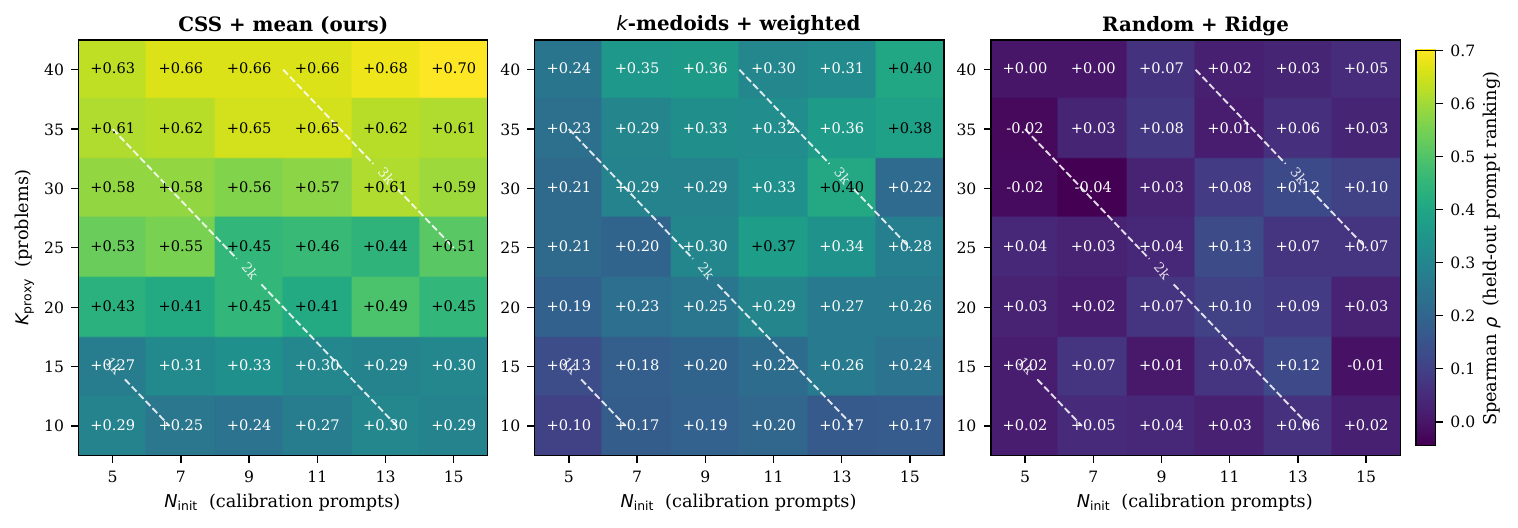}
  \caption{Proxy-benchmark ranking quality as a function of $(N_{\mathrm{init}}, K_{\mathrm{proxy}})$, where $N_{\mathrm{init}}$ is the number of calibration prompts scored on the full discovery set and $K_{\mathrm{proxy}}$ is the size of the proxy problem subset used during evolution. Each cell is the mean Spearman rank correlation between proxy and full-set scores on held-out prompts, averaged over 60 random splits of a 24-prompt $\times$ 150-problem pool. Dashed white contours mark equal-budget reallocations under total LLM calls $= 150 \cdot N_{\mathrm{init}} + 50 \cdot K_{\mathrm{proxy}}$ (assuming 50 evolution iterations). The greedy column-subset method (left) preserves rankings substantially better than $k$-medoids + cluster-size-weighted mean (middle) and random-subset + ridge regression (right) at every iso-cost contour. Ridge collapses because its $K_{\mathrm{proxy}}$ free weights are underdetermined by only $N_{\mathrm{init}} \le 15$ samples.}
  \label{fig:budget-tradeoff}
\end{figure}

\Cref{fig:budget-tradeoff} shows that CSS+mean dominates at every cell, peaking at $\rho = 0.70$ versus $0.40$ for the $k$-medoids baseline and $0.07$ for random-subset + ridge; the gap holds along every iso-cost contour. At a fixed budget of ${\sim}2{,}500$ LLM calls---e.g., $(N_{\mathrm{init}}, K_{\mathrm{proxy}}) = (5, 35)$---CSS+mean attains $\rho \approx 0.61$ versus $0.31$ for $k$-medoids and $0.04$ for ridge. The ridge baseline collapses because its $K_{\mathrm{proxy}}$ free weights are underdetermined by only $N_{\mathrm{init}} \le 15$ training samples; the $k$-medoids baseline has the right form but the wrong objective---it picks representative problems without optimizing for the candidate-ordering signal that evolution actually consumes, which is exactly the gap CSS+mean's rank-faithfulness term targets.

\section{Archive composition for the diversity ablation}
\label{app:archive-composition}

\begin{figure}[!htbp]
  \centering
  \includegraphics[width=\textwidth]{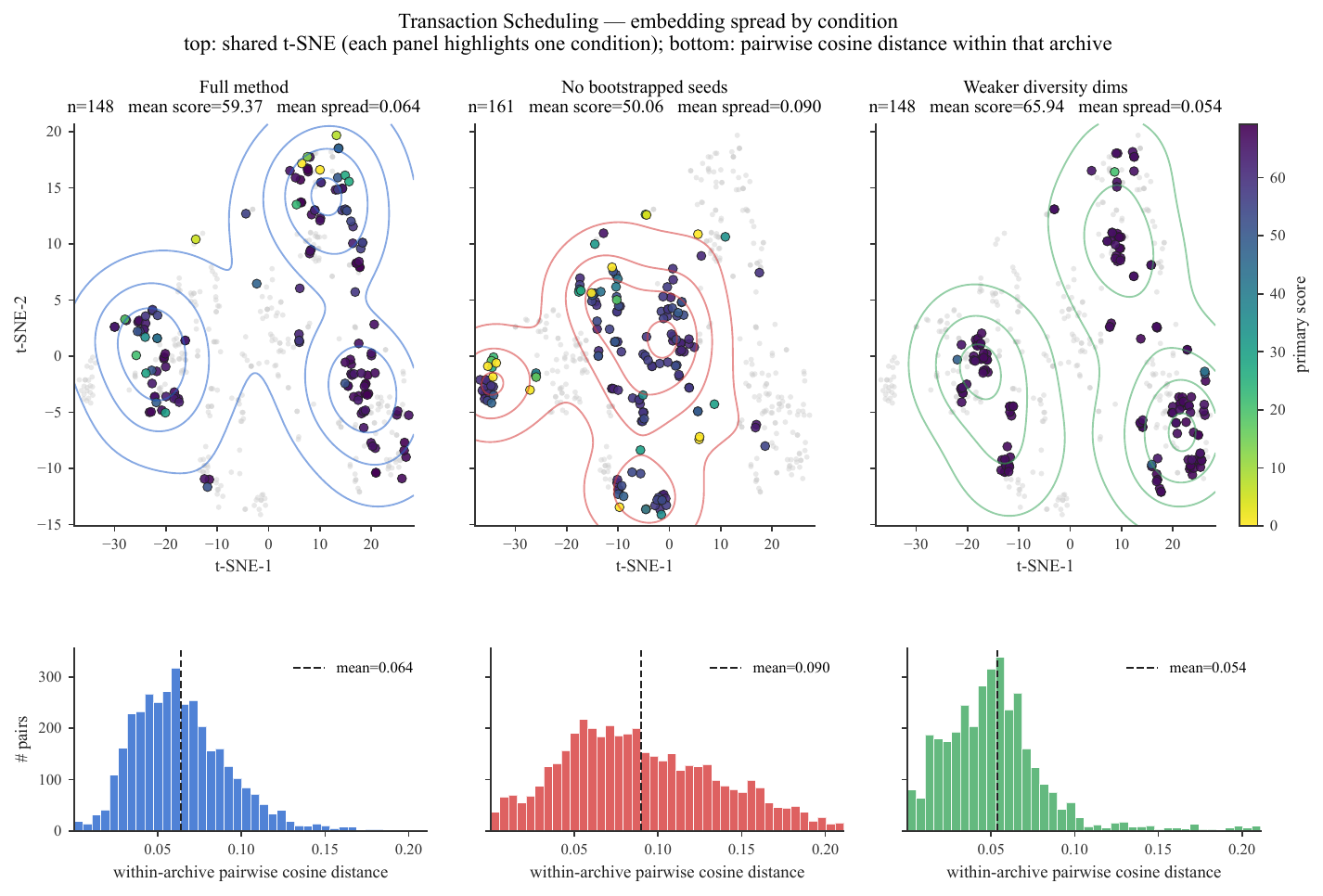}
  \caption{Transaction Scheduling --- final-archive composition for the three diversity-ablation conditions. Top: shared t-SNE of all elites coloured by primary score (darker = higher); each panel highlights one condition with KDE contours, with the other conditions shown faintly in gray for spatial reference. Bottom: distribution of within-archive pairwise cosine distances (computed within each seed and pooled across three seeds; no cross-seed pairs).}
  \label{fig:archive-embeddings-txn}
\end{figure}

\begin{figure}[!htbp]
  \centering
  \includegraphics[width=\textwidth]{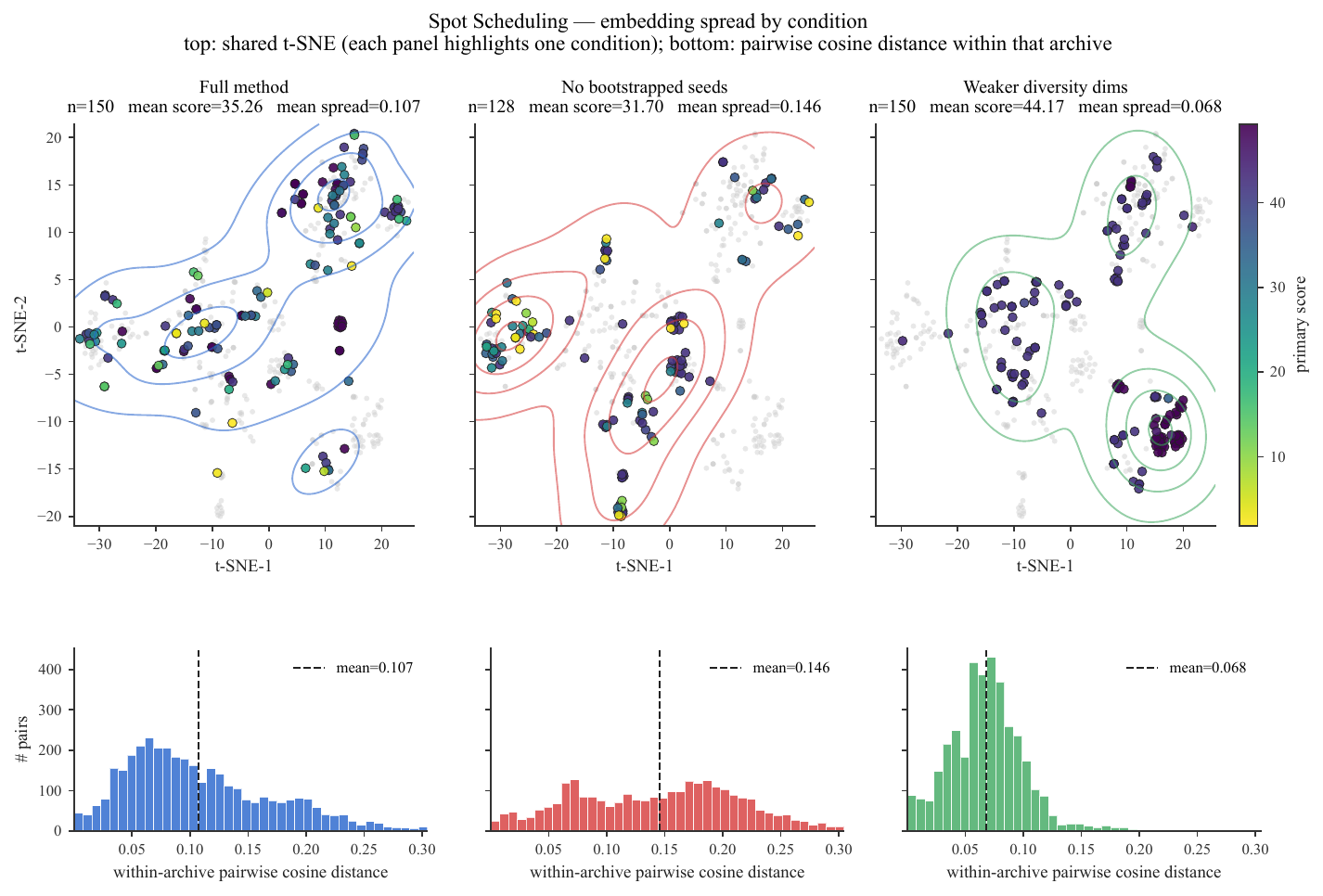}
  \caption{Spot Scheduling --- final-archive composition for the three diversity-ablation conditions. Plotting conventions as in \Cref{fig:archive-embeddings-txn}.}
  \label{fig:archive-embeddings-cbl}
\end{figure}

We embed every elite from the diversity ablation (\Cref{fig:diversity-ablation}) with OpenAI's \texttt{text-embedding-3-small} and inspect the resulting $1{,}536$-dimensional archives. \Cref{fig:archive-embeddings-txn,fig:archive-embeddings-cbl} show, per condition: the joint t-SNE projection coloured by primary score (top), and the distribution of within-archive pairwise cosine distances pooled across the three seeds (bottom). \emph{No bootstrapped seeds} produces the widest spread on both problems due to its uniform initialization (mean within-archive cosine distance $0.090$ on Transaction Scheduling and $0.146$ on Spot Scheduling) but populates that spread with low-scoring cells; \emph{weaker dimensions} produces the tightest archive on both problems ($0.054$ and $0.068$) showing high level of convergence; the full method sits between the tw, showing a mix of diverse solutions that are also high scoring.

To check whether tightness corresponds to fewer algorithmic families, we cluster each problem's embeddings with k-means ($k{=}5$) and label each cluster from three nearest-to-centroid representatives using GPT-4.1-mini. \Cref{tab:cbl-families} reports the result for Spot Scheduling: weaker dimensions concentrate $35\%$ of their elites in a single ``deadline-aware restart-overhead cautious'' family (C2) and $42\%$ in C1, while the full method's elites distribute across four of the five families and reach peaks (C4 best $50.4$) that the collapsed weak-dims archive does not attain (best $49.3$). On Transaction Scheduling all five clusters are sub-variants of the conflict-aware-greedy family---consistent with the body observation that ``every elite across the nine runs sits in the conflict-aware-greedy family''---and the family distribution is largely insensitive to the dimension choice.

\begin{table}[!htbp]
  \centering
  \small
  \setlength{\tabcolsep}{3pt}
  \caption{Algorithmic-family counts in the Spot Scheduling final archives. Clusters are obtained by k-means ($k{=}5$) on \texttt{text-embedding-3-small} embeddings of every elite across three seeds per condition; labels are produced by feeding three nearest-to-centroid representatives to GPT-4.1-mini. Bold marks each condition's modal cluster.}
  \label{tab:cbl-families}
  \begin{tabular}{lrrrrrr}
  \toprule
  Cluster (algorithmic family) & a Full & b No-boot & c Weak-dims & $n$ & mean & best \\
  \midrule
  C4 Preemption-aware deadline-safe SPOT fallback   & \textbf{75} & 43          & 33          & 151 & 38.4 & 50.4 \\
  C1 Risk-aware dynamic fallback scheduling          & 45          & 34          & \textbf{63} & 142 & 37.3 & 50.7 \\
  C2 Deadline-aware restart-overhead cautious sched. & 12          & 3           & 52          & 67  & 44.5 & 49.3 \\
  C3 Probabilistic entropy-based adaptive sched.     & 17          & \textbf{34} & 2           & 53  & 27.6 & 49.6 \\
  C0 Restart-overhead estimation variants            & 1           & 14          & 0           & 15  & 29.3 & 42.2 \\
  \midrule
  Total elites                                       & 150         & 128         & 150         & 428 &      &      \\
  \bottomrule
  \end{tabular}
\end{table}


\end{document}